%% file: main.tex
\documentclass[letterpaper]{article} 
\usepackage{aaai25}  
\usepackage{times}  
\usepackage{helvet}  
\usepackage{courier}  
\usepackage[hyphens]{url}  
\usepackage{graphicx} 
\usepackage{natbib}  
\usepackage{caption} 
\usepackage{algorithm}
\usepackage{algorithmic}
\usepackage{xfrac}
\usepackage{mathtools}
\usepackage{booktabs}       
\usepackage{mathabx}
\usepackage{subfig}
\usepackage{xcolor}         
\usepackage{multirow}
\usepackage{anyfontsize}

\usepackage{newfloat}
\usepackage{listings}
\DeclareCaptionStyle{ruled}{labelfont=normalfont,labelsep=colon,strut=off} 
\floatstyle{ruled}
\newfloat{listing}{tb}{lst}{}
\floatname{listing}{Listing}
\newcommand{\norm}[1]{\left \lVert #1 \right \rVert}

\title{Qua$^2$SeDiMo: Quantifiable Quantization Sensitivity of Diffusion Models}
\author{
    Keith G. Mills\textsuperscript{\rm 1,\rm 2}, 
    Mohammad Salameh\textsuperscript{\rm 2}, 
    Ruichen Chen\textsuperscript{\rm 1}, 
    Negar Hassanpour\textsuperscript{\rm 2}, 
    Wei Lu\textsuperscript{\rm 3}, 
    Di Niu\textsuperscript{\rm 1} 
}
\affiliations{
    \textsuperscript{\rm 1}Department of Electrical and Computer Engineering, University of Alberta\\
    \textsuperscript{\rm 2}Huawei Technologies, Edmonton, Alberta, Canada\\
    \textsuperscript{\rm 3}Huawei Kirin Solution, Shanghai, China\\
    \{kgmills, ruichen1, dniu\}@ualberta.ca\\
    \{mohammad.salameh, negar.hassanpour2\}@huawei.com\\
    robin.luwei@hisilicon.com\\
}

\usepackage{bibentry}

\begin{document}

\maketitle

\begin{abstract}
\input{src/abstract}
\end{abstract}

%
 \begin{links}
     \link{Project}{https://kgmills.github.io/projects/qua2sedimo/}
 \end{links}

\input{src/intro}
\input{src/related}
\input{src/method}
\input{src/results}

\input{src/conclusion}

\clearpage

\section*{Acknowledgements}
This work is partially funded by an Alberta Innovates Graduate Student Scholarship (AIGSS). Alberta Innovates is an organization that expands the horizon of possibilites to solve today's challenges anc reate a healthier and more prosperous future for Alberta and the world. 

\bibliography{aaai25}

\clearpage

\input{src/appendix}

\end{document}

%% file: src/abstract.tex
Diffusion Models (DM) have democratized AI image generation through an iterative denoising process. Quantization is a major technique to alleviate the inference cost and reduce the size of DM denoiser networks. However, as denoisers evolve from variants of convolutional U-Nets toward newer Transformer architectures, it is of growing importance to understand the quantization sensitivity of different weight layers, operations and architecture types to performance. In this work, we address this challenge with Qua$^2$SeDiMo, a mixed-precision Post-Training Quantization framework that generates explainable insights on the cost-effectiveness of various model weight quantization methods for different denoiser operation types and block structures. We leverage these insights to make high-quality mixed-precision quantization decisions for a myriad of diffusion models ranging from foundational U-Nets to state-of-the-art Transformers. As a result, Qua$^2$SeDiMo can construct 3.4-bit, 3.9-bit, 3.65-bit and 3.7-bit weight quantization on PixArt-$\alpha$, PixArt-$\Sigma$, Hunyuan-DiT and SDXL, respectively. We further pair our weight-quantization configurations with 6-bit activation quantization and outperform existing approaches in terms of quantitative metrics and generative image quality.

%% file: src/intro.tex
\section{Introduction}
\label{sec:introduction}

Diffusion Models (DM)~\cite{sauer2024fast} have become the state of the art in image synthesis. However, at the core of every DM is a large denoiser network, e.g., a U-Net or Diffusion Transformer. The denoiser performs multiple rounds of inference, thus imposing a significant computational burden on the generative process.

One effective method for reducing this burden is quantization~\cite{du2024model} which reduces the bit precision of weights and activations. TFMQ-DM~\cite{huang2023tfmq}, a state-of-the-art DM Post-Training Quantization (PTQ) approach, carefully quantizes weight layers associated with time-step inputs to ensure accurate image generation. Q-Diffusion~\cite{li2023q} split the weight layers associated with long residual connections to compensate for bimodal activation distributions and has been integrated into Nvidia's TensorRT framework~\cite{nvidia2024qdiff}. Additionally, ViDiT-Q adopt Large Language Model (LLM) quantization techniques~\cite{xiao2023smoothquant} to compress newer Diffusion Transformers (DiT)~\cite{peebles2023scalable} like the PixArt models~\cite{chen2024pixartAlpha, chen2024pixartsigma}. In order to preserve generation quality, each of these techniques employs a calibration set to perform gradient-based calibration for weight quantization. However, till now existing methods still struggle to quantize weight precision below 4-bits (W4) in diffusion models without severely degrading the image generation quality.

To achieve low-bit quantization, mixed-precision quantization has recently been explored for LLMs, although not for DMs yet, which aims to differentiate the bit precision applied to different weights. Talaria~\cite{hohman2024talaria} is a tool developed by Apple to visualize the impact of different compression techniques applied to different model layers on hardware metrics and latency, which however cannot assess the same impact on task performance. OWQ~\cite{lee2024owq} attempts to identify weight column vectors that can generate outlier activations in LLMs, while PB-LLM~\cite{shang2023pb} measures the salience of individual weights. Such outlier or salience information is then used to apply different quantization configurations and bit precisions across weights in an LLM. Although these techniques rely on the Hessian of model weights to identify sensitive weights, the weight saliency is computed by heuristics and is not directly derived to associate with task performance.  Another limitation is that as originally designed for LLMs, the granularity adopted is on a fine-grained weight (or weight column) level, rather than on an operator (layer) level or block level.  However, such generalizable per-operator or per-model insights are valuable for DMs, which involve a diverse range of model types, e.g., various types of U-Nets and DiTs, as well as a wider range of operation types than in LLMs. These insights, if available, will not only help differentiating quantization method and configuration selection per operation, but also help identify specific operation types, e.g., time-step embeddings or skip-connections that can greatly affect end-to-end performance when improperly quantized.

\begin{figure*}[t]
    \centering
    \includegraphics[width=6.9in]{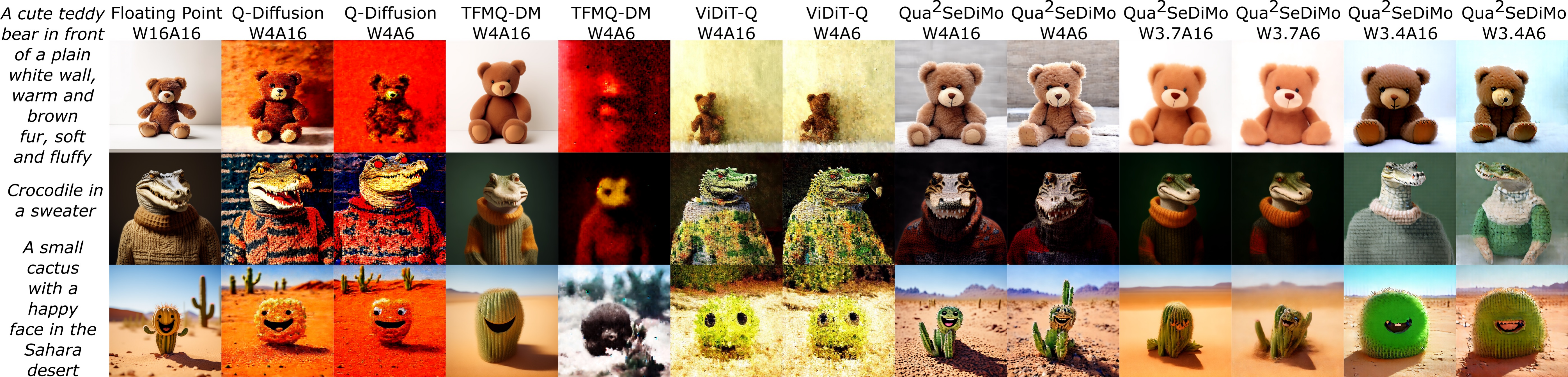}
    \caption{Example $512^2$ images generated using PixArt-$\alpha$. We compare images from the full precision model to those generated by a quantized denoiser using different PTQ techniques. Specifically, we compare Q-Diffusion, TFMQ-DM and ViDiT-Q at W4 precision to three configurations built by Qua$^2$SeDiMo - W4, W3.7 and W3.4 - with and without 6-bit activation quantization.}
    \label{fig:alpha_images}
\end{figure*}

In this paper, we propose Qua$^2$SeDiMo (pronounced kwa-see-dee-mo), short for \textbf{Qua}ntifiable \textbf{Qua}ntization \textbf{Se}nsitivity of \textbf{Di}ffusion \textbf{Mo}dels, a framework for discovering PTQ sensitivity of components in various types of DMs to user-defined end-to-end objectives, including task performance and model complexity. Qua$^2$SeDiMo can identify the individual weights, operation types and block structures that disproportionately impact end-to-end image generation performance when improperly quantized as well as higher-level insights regarding the preference of model and operation types for different quantization schemes and configurations. Furthermore, we combine the algorithm-discovered insights to construct mixed precision, sub 4-bit weight quantization configurations that facilitate high-quality image synthesis, as illustrated by Figure~\ref{fig:alpha_images} for PTQ performed over a contemporary DiT model, PixArt-$\alpha$. Our contributions are as follows:

First, unlike previous approaches that use Hessian and other proxies to identify sensitive weights, we propose a method to correlate the quantization method and bit precision of every layer (operation) directly to end-to-end network metrics such as model size or task performance. This is challenging because denoisers in DMs contain hundreds of layers, resulting in exponentially many quantization configuration combinations in the whole network. Moreover, DMs require costly computation to evaluate even with PTQ. However, our method can learn to assign the optimal configuration to each layer by evaluating less than 500 sampled quantization configurations. Qua$^2$SeDiMo achieves this by representing denoiser architectures as graphs, then leveraging an optimization-based GNN explanation method to attribute graph-level performance to individual layers as well as larger block structures like self-attention and temporal embedding layers.

Second, our insights reveal which specific model layers, blocks and quantization methods make sub 4-bit PTQ difficult. Specifically, we find that while U-Nets have a preference for uniform, scale-based quantization~\cite{nahshan2021loss}, DiT models prefer cluster-based~\cite{han2015deep} methods. Additionally, we show that the ResNet blocks in U-Nets are more sensitive than DiT Transformer blocks to quantization, requiring higher bit precision to maintain end-to-end performance and image quality. We also find that the final output layer of DiT models are more sensitive to quantization than their U-Net counterparts.

Third, we construct efficient, mixed-precision weight quantization configurations that generate high-fidelity images. Specifically, we achieve 3.4, 3.9, 3.65, 3.7 and 3.5-bit PTQ on PixArt-$\alpha$, PixArt-$\Sigma$, Hunyuan-DiT~\cite{li2024hunyuandit}, SDXL and DiT-XL/2, respectively, without requiring a calibration dataset. Finally, we pair our weight-quantization with activation quantization, outperforming existing techniques like Q-Diffusion, TFMQ-DM~\cite{huang2023tfmq} and ViDiT-Q~\cite{zhao2024vidit} in terms of visual quality, FID and CLIP scores. 

%% file: src/related.tex
\section{Related Work}
\label{sec:related}

Diffusion models~\cite{sohl2015deep, jiang2024frap} are a class of generative models that have been successfully adopted to generate high-fidelity visual content~\cite{sauer2024fast}. DMs utilize a progressive denoising process to achieve state-of-the-art image generation. Mainstream approaches for high-resolution image generation leverage the latent space of a Variational Auto-Encoder (VAE)~\cite{kingma2013auto} by placing a large denoiser network between the VAE encoder and decoder. Foundational text-to-image (T2I) DMs like SDv1.5 and SDXL~\cite{podell2023sdxl} adopt a hierarchical U-Net-based denoiser architecture that blends Convolutional and Transformer block structures. However, more recent DMs like DiT and SD3~\cite{esser2024scaling} use non-hierarchical patch-based architectures based on Vision Transformers~\cite{frumkin2023jumping}. Our proposed method, Qua$^2$SeDiMo, is architecture agnostic, so we consider both architecture styles in this work.

The iterative denoising process makes DMs slow. Such a limitation is addressed through model optimization techniques, such as quantization~\cite{gholami2022survey}. Quantization reduces the bit precision of neural network weights and activation from $\ge$16-bit Floating Point (FP) formats to $\le$8-bit Integer (INT)/FP~\cite{shen2024efficient} formats. There are two broad approaches: Post-Training Quantization (PTQ)~\cite{lin2023awq, lee2024owq} can be applied to pre-trained model weights, while Quantization-Aware Training (QAT)~\cite{sui2024bitsfusion} trains or fine-tunes weights in an end-to-end manner using Straight-Through Estimators~\cite{huh2023straightening} to preserve gradient flow. PTQ is generally computationally inexpensive relative to QAT, though some approaches~\cite{nagel2020up, li2021brecq} rely on a calibration dataset of unlabeled sample data. PTQ tends to encounter issues below 4-bit precision~\cite{frumkin2023jumping, krishnamoorthi2018quantizing} while QAT can quantize Large Language Models (LLM)~\cite{touvron2023llama} weights to a very low precision of 1.58-bits~\cite{ma2024era}. By contrast, this work achieves sub 4-bit mixed precision PTQ for DM weights without requiring a calibration set, after which activation quantization can be applied with minimal performance loss.

Several PTQ~\cite{he2024ptqd, zhao2025mixdq} and QAT~\cite{wang2024quest} approaches exist for DMs. The earliest publications, PTQ4DM~\cite{shang2023post} and Q-Diffusion~\cite{li2023q} emphasize the importance of carefully sampling a calibration dataset to quantize denoiser activations properly. TDQ~\cite{so2024temporal} uses an auxiliary model to generate activation quantization parameters for different denoising steps while QDiffBench~\cite{tang2023post} relaxes activation bit precision at the start and end of the denoising process. Most approaches study the impact of quantization on the denoising process, while a few study the quantization sensitivity of denoiser weight types. For example, Q-Diffusion and TFMQ-DM proposed novel techniques to quantize the long residual connections and time-step embedding layers, respectively. However, these insights are specific to U-Net-based denoisers. By contrast, our work extends this discussion by studying the quantization sensitivity of all weight types and positions while introducing a generalizable method applicable to any denoiser architecture. 

%% file: src/method.tex
\section{Background}
\label{sec:background}

We provide a briefing on several integer-based weight quantization methods, including how they are performed and impact on DM generative performance. 

Given a tensor $W_{FP}$ with precision $N_{FP}$, we \textit{quantize} it into $W_Q$ with precision $N_Q$, thus reducing the tensor size by a factor of $\sfrac{N_{Q}}{N_{FP}}$. At inference time, we \textit{dequantize} $W_Q$ into $W_{DQ}$ with precision $N_{FP}$. Although $W_{DQ}$ has 
the same precision as $W_{FP}$, quantization introduces an error $\epsilon = \norm{W_{FP} - W_{DQ}}_p$, where $p\ge2$; we refer interested readers to \citet{nahshan2021loss} for further discussion on $p$.

One method to perform quantization is by applying $K$-Means clustering~\cite{han2015deep} to $W_{FP}$. Specifically, we can cluster across the entire tensor or each output channel $c_{out}$ of $W_{FP}$ separately. In either case, $W_Q$ is a matrix of indices corresponding to $K=2^{N_Q}$ cluster centroids of precision $N_{FP}$. However, as $W_{DQ}$ is created by substituting the indices with their corresponding centroids, dequantization is slower and not as hardware-friendly as other methods~\cite{jacob2018quantization}. Additionally, this form of quantization incurs a high FP overhead as centroids are kept in $N_{FP}$-bit precision. We refer readers to the supplementary for computation of the FP overhead.

In contrast to the costly $K$-Means, another popular PTQ method is Uniform Affine Quantization (UAQ)~\cite{krishnamoorthi2018quantizing}, which involves computing a scale $\Delta$, 

\begin{equation}
    \centering
    \label{eq:uaq_delta}
    \Delta = \dfrac{\texttt{max}(|W_{FP}|)}{2^{N_Q-1}-1}. 
\end{equation}

Then, tensor quantization is performed by rescaling and clamping $W_{FP}$ as follows, 

\begin{equation}
    \centering
    \label{eq:uaq_wq}
    W_Q = \texttt{clamp}(\lfloor\dfrac{W_{FP}}{\Delta}\rceil, -2^{N_Q-1}+1, 2^{N_Q-1}-1),
\end{equation}
where $\lfloor\cdot\rceil$ is the rounding operation. Note that for simplicity, we assume $W_{FP}$ is symmetric at 0~\cite{xiao2023smoothquant}. The computation is similar when factoring in a zero-point $z$ for asymmetric quantization~\cite{jacob2018quantization}. UAQ places the tensor values into $2^{N}$ evenly-spaced bins of width $\Delta$. UAQ is performed per output channel $c_{out}$ when performing weight quantization. UAQ has two key advantages over $K$-Means: First, the FP overhead is smaller as we only need to save $\Delta$ (and $z$, if applicable) as FP scalars. Second, UAQ dequantization is simple multiplication $W_{DQ} = \Delta W_Q$ which is very efficient on modern hardware using kernel fusion~\cite{lin2023awq}, making it the preferred method for deployment on edge devices.

\begin{figure}[t]
    \centering
    \includegraphics[width=3.2in]{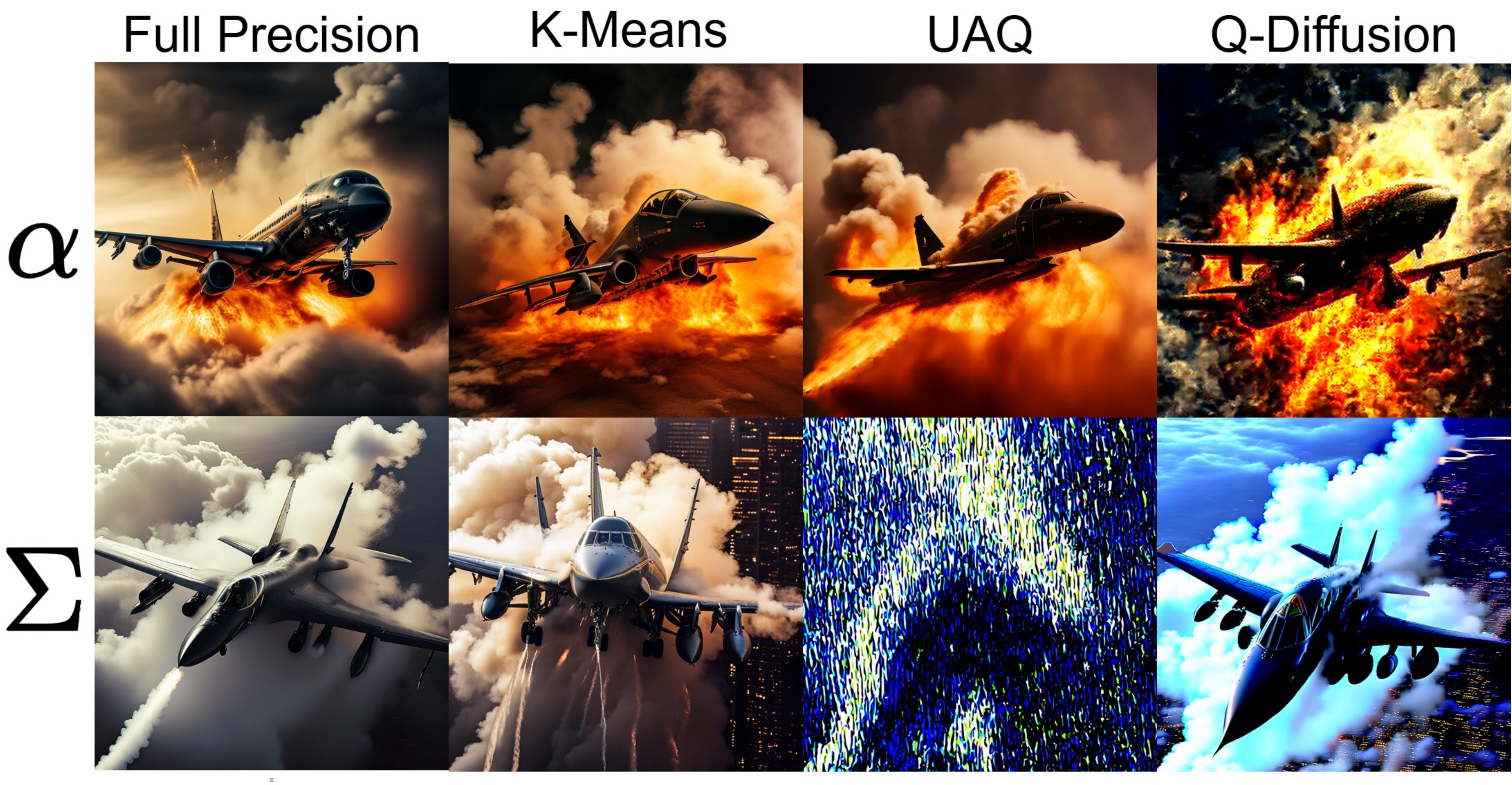}
    \caption{PixArt-$\alpha$/$\Sigma$ images at FP16 precision and quantized to W4A16 by $K$-Means, UAQ and Q-Diffusion. COCO prompt: `A jet with smoke pouring from its wings'.} 
    \label{fig:w4a16}
\end{figure}

\begin{table}[t]
    \centering
    \scalebox{0.9}{
    \begin{tabular}{lcccc} \toprule
    \textbf{Quant. Method} & \textbf{$\alpha$-W4} & \textbf{$\alpha$-W3} & \textbf{$\Sigma$-W4} & \textbf{$\Sigma$-W3} \\ \midrule
    \textbf{K-Means} & 32.99 & 80.65 & 38.13 & 133.87 \\ 
    \textbf{UAQ w/Eq.~\ref{eq:alpha}} & 32.74 & 77.06 & 143.25 & 447.11 \\ 
    \textbf{Q-Diffusion} & 40.04 & 264.51 & 43.14 & 366.45 \\  \bottomrule
    \end{tabular}
    }
    \caption{FID score for PixArt-$\alpha$/$\Sigma$ generating 10k images using MS-COCO prompts under different weight quantization (W\{3, 4\}A16) configurations. 
    Lower FID is better; FID of the FP model is \textbf{34.05} and \textbf{36.94} for $\alpha$ and $\Sigma$, respectively.}
    \label{tab:fid_1k}
\end{table}

However, note that Eq.~\ref{eq:uaq_delta} is deterministic and may not be optimal. One way to address this problem is to reduce $\Delta$, 

\begin{equation}
    \centering
    \label{eq:alpha}
    \Delta_{\alpha} = \dfrac{\texttt{max}(|W_{FP}|)\times(1-(0.01\alpha))}{2^{N_Q-1}-1}.
\end{equation}
where $\alpha \in [0, 100)$ is selected to minimize the $L_p$ loss:  
\begin{equation}
    \centering
    \label{eq:alpha_loss}
    \min_{\alpha}\norm{W_{FP}-\Delta_{\alpha}W_{Q}}_p.
\end{equation}

While it is straightforward to apply Eqs.~\ref{eq:alpha} and \ref{eq:alpha_loss} to individual operators, advanced PTQ methods like AdaRound~\cite{nagel2020up} and BRECQ~\cite{li2021brecq} use higher-order loss information to refine $\Delta$. These methods are the basis of advanced DM PTQ schemes like Q-Diffusion. However, they require a calibration set to function, which is not required by $K$-Means or UAQ when quantizing weights. 

As Fig.~\ref{fig:w4a16} shows, using any of these methods to quantize denoiser weights down to $N_Q=4$ produces images that are similar in detail and/or structure to ones generated by the FP model. To quantify the performance, we generate 10k images using MS-COCO~\cite{coco} prompts and measure the Fr\'echet Inception Distance (FID)~\cite{heusel2017gans} using the validation set. As Table~\ref{tab:fid_1k} shows, all three methods achieve comparable or even lower FID relative to the FP16 model (not unheard of for DM PTQ~\cite{shang2023post, li2023q, he2024ptqd, huang2023tfmq}) at 4-bit precision. However, further weight quantization to $N_Q = 3$ yields a sharp rise in FID. We hypothesize that this occurs because PTQ methods quantize every weight to the same precision. Thus, we strike a balance by generating 4 and 3-bit mixed precision weight PTQ configurations.

\section{Methodology}
\label{sec:methodology}

In this section, we elaborate on our search space and describe how to cast a DM denoiser as a graph. We then measure the quantization sensitivity of weight layers and block structures by a GNN explanation method that correlated end-to-end performance with operations and blocks.

\begin{figure}
    \centering
    \includegraphics[width=3.2in]{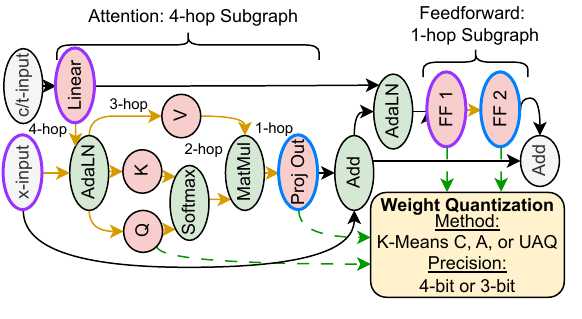}
    \caption{Induced DiT subgraphs. Attention weights (red) are captured in a 4-hop subgraph rooted at `Proj Out'. The feedforward module is a 1-hop subgraph rooted at `FF 2'. Yellow box: Each weight layer can be quantized using three methods and two bit-precision levels.}
    \label{fig:subgraphs}
\end{figure}

We form a search space for each denoiser by varying the bit precision and quantization method applied to each weight layer. The yellow box in Fig.~\ref{fig:subgraphs} enumerates the available choices. Specifically, we consider two bit-precisions $N_Q = \{3, 4\}$ and three quantization methods: \hbox{$K$-Means C}, \hbox{$K$-Means A} and UAQ. \hbox{$K$-Means C} quantizes each applies output channel $c_{out}$ separately while \hbox{$K$-Means A} applies clustering to the entire tensor for smaller FP overhead. UAQ utilizes an optimal $\alpha$ value, predetermined using a simple grid search of 10 choices $\alpha \in [0, 10,..., 80, 90]$ per layer.

In sum, this provides us with 6 quantization choices per weight layer and a total search space size of $6^{\#W}$ where $\#W$ is the number of quantizable weight layers across the entire denoiser architecture. We refer to a denoiser architecture where all weight layer nodes have been assigned a specific bit precision and quantization method as a \textit{quantization configuration}. We can sample various configurations from the search space, apply them to the original FP DM denoiser network, generate images, and then measure end-to-end statistics like FID and average bit precision. Next, we describe how to exploit the properties of graph structures to extract meaningful insights about the denoiser search space.

\subsection{Operation-Level Sensitivity via Graphs}
\label{sec:autobuild}

We represent denoiser architectures as Directed Acyclic Graphs (DAG)~\cite{mills2023gennape} where nodes represent weight layers, e.g., \texttt{nn.Linear} or \texttt{nn.Conv2d}, and other operations like `Add', while the edges model the forward-pass information flow. We provide an example illustration in Figure~\ref{fig:subgraphs} where red nodes correspond to quantizable weights. We encode the quantization method, bit precision, operation type (e.g., attention `query' linear layer) and positional information like Transformer block index as node features. This encoding allows us to extract quantifiable insights on the sensitivity of denoiser architectures by identifying the operation types, block structures, positions and quantization methods that contribute to high end-to-end performance and low average bit precision. To achieve this, we introduce the following explanation method for Graph Neural Network (GNN)~\cite{brody2022how, Fey/Lenssen/2019} regressors:

Let $\mathcal{G}$ be a denoiser graph with specific quantization configuration, annotated with ground-truth label $y_\mathcal{G}$, e.g., negative FID $y_\mathcal{G}=-FID_\mathcal{G}$. $\mathcal{G}$ contains a node set $\mathcal{V}_\mathcal{G}$, whose features describe the quantization settings for each weight layer, and edge set $\mathcal{E}_\mathcal{G}$. We can then use a GNN to learn to predict $y_\mathcal{G}$ given $\mathcal{G}$. A GNN contains $m \in [0, M]$ layers: an initial embedding layer followed by $M$ message passing layers, each of which produces an embedding $h_{v_i}^m$ for every node $v_i \in \mathcal{V}_\mathcal{G}$. Node embeddings from a given GNN layer can be aggregated to form a vector embedding for the graph, e.g., by averaging them,

\begin{equation}
    \centering
    \label{eq:aggr}
    h_\mathcal{G}^m = \dfrac{1}{|\mathcal{V}_\mathcal{G}|}\sum_{v\in\mathcal{V}_\mathcal{G}}h_v^m. 
\end{equation}

We can then apply a simple MLP to the graph embedding to make a prediction $y'_\mathcal{G} = \texttt{MLP}(h_\mathcal{G}^M)$. A simple GNN can learn by minimizing a loss, e.g., mean-squared error $\norm{y_\mathcal{G}-y_\mathcal{G}'}_2$, which we denote as $\mathcal{L}_{orig}$. This formulation is a typical black box model and while it can estimate $y_\mathcal{G}$, it cannot extract quantifiable insights. Instead, this is accomplished by incorporating an additional loss term: 

\begin{equation}
    \centering
    \label{eq:ab_loss}
    \mathcal{L} = \mathcal{L}_{orig}(y_\mathcal{G}, y_\mathcal{G}') + \dfrac{1}{M+1}\sum_{m=0}^{M}\mathcal{L}_{rank}(y_\mathcal{G}, \norm{h_\mathcal{G}^m}_1),
\end{equation}
where $\mathcal{L}_{rank}$ is a ranking loss that directly interfaces with the graph embeddings $h_\mathcal{G}^m$ from each GNN layer. The exact choice of $\mathcal{L}_{rank}$ is important. A straight-forward idea is to choose the differentiable spearman $\rho$ loss that \citet{blondel2020fast} provide, in order to maximize the Spearman Rank Correlation Coefficient (SRCC). However, SRCC assigns equal importance to the predicted rank of every entry considered, weighing entries that \textit{minimize} and \textit{maximize} the ground-truth equally. In contrast, depending on how we compute $y_\mathcal{G}$, our goal is to extract insights from the graphs that explicitly maximize $y_\mathcal{G}$. Therefore, one alternative is to maximize the Normalized Discounted Cumulative Gain (NDCG), an Information Retrieval metric that prioritizes the correct ranking of high-relevance (i.e., high $y_\mathcal{G}$) samples, by implementing the LambdaRank~\cite{burges2010ranknet} loss.

Regardless of the choice of $\mathcal{L}_{rank}$, the intuition behind our approach is to compress $h_\mathcal{G}^m$ into its scalar L1 norm and associate it with the ground-truth $y_\mathcal{G}$. Then, because $h_\mathcal{G}^m$ is computed by averaging all node embeddings per Eq.~\ref{eq:aggr}, the GNN is forced to learn which nodes contribute or detract from $y_\mathcal{G}$. As such, we are able to treat the scalar norm of the node embedding $\norm{h_{v_i}^m}_1$ as a numerical score where high $\norm{h_{v_i}^m}_1$ correspond to higher $y_\mathcal{G}$.

Finally, using this setup we can construct highly desirable quantization configurations. Assume we train a predictor using Eq.~\ref{eq:ab_loss} where $y_\mathcal{G}=-FID_\mathcal{G}$. We can select the optimal bit precision and quantization method for every weight layer node simply by iterating across all possible combinations (i.e., 6 per node according to Fig.~\ref{fig:subgraphs}) and selecting the configuration that produces the highest score $\norm{h_{v_i}^0}_1$.

\subsection{Block-level Quantization Sensitivity}
\label{sec:quant_sen_subgraphs}

While we have shown how Eq.~\ref{eq:ab_loss} produces sensitivity scores for individual weight nodes, it is non-trivial to extend this idea to larger denoiser components, e.g., ResNet blocks or time-step embedding modules. To do this, we model these structures as subgraphs contained within the overall denoiser graph. Each subgraph contains a root node corresponding to a single operation. The root only aggregates information (e.g., quantization method and precision features) from the other nodes in its subgraph, allowing us to interpret its score as representative of the entire subgraph block structure. 

As a practical example, Figure~\ref{fig:subgraphs} provides an illustration where a DiT-XL/2 Transformer block is split into attention and feedforward subgraphs, rooted at the `Proj Out' and `FF 2' weight nodes, respectively. Therefore, we cast $\norm{h^4_{Proj Out}}_1$ and $\norm{h_{FF 2}^1}_1$ as the sensitivity scores for the attention and feedforward modules, respectively. Further, we can then construct high-quality quantization configuration by looping over all quantization method and bit precision choices for each weight layer node in the subgraph and selecting the option that yields the greatest score.

Note that this schema contains several design choices and details about the block structures we cast as subgraphs and which weights should be chosen as roots. Generally, we root our subgraphs at the last weighted layer of the block structure, though there are some exceptions and we provide extensive details in the supplementary.

Further, it should be noted that we are able to quantify the sensitivity of large block structures by exploiting the message passing properties of GNNs. Formally, given an arbitrary node $v_i \in \mathcal{V}_\mathcal{G}$, a single GNN layer will propagate latent embeddings $h_{v_j}$ from all nodes in the immediate 1-hop neighborhood $\mathcal{N}(v_i) = \{v_j \in \mathcal{V}_{\mathcal{G}} | (v_j, v_i) \in \mathcal{E}_\mathcal{G}\}$ into the embedding of $v_i$, $h_{v_i}$. Applying another GNN layer will further propagate information from all nodes in the 2-hop neighborhood of $v_i$ into $h_{v_i}$. 

We define the $m$-hop neighborhood $\mathcal{N}^m(v_i) \subseteq \mathcal{V}_\mathcal{G}$ as $\mathcal{N}^m(v_i) = \{v_j \in \mathcal{V}_\mathcal{G} | \langle v_j, v_i \rangle \leq m\}$, where $\langle v_j, v_i \rangle$ is the length of the shortest path between $v_j$ and $v_i$. By induction, applying $m>0$ GNN layers will aggregate information from all nodes in $\mathcal{N}^m(v_i)$ into $h_{v_i}^m$. As such, we extend the meaning of $h_{v_i}^m$ from not simply the embedding of node $v_i$, but as the embedding of the subgraph containing all nodes in $\mathcal{N}^m(v_i)$ that is rooted at $v_i$. Likewise, we can now interpret $\norm{h_{v_i}^m}_1$ as the quantifiable score of this subgraph.  

%% file: src/results.tex
\section{Experimental Results and Discussion}
\label{sec:results}

\begin{table}
    \centering
    \scalebox{0.9}{
    \begin{tabular}{lcccc} \toprule
    \textbf{Denoiser} & \textbf{FID} & \textbf{$\#W$} & \textbf{\#Configs} & \textbf{FID Range} \\ \midrule 
    \textbf{PixArt-$\alpha$} & 99.67 & 287 & 372 & [92.29, 507.87] \\ 
    \textbf{PixArt-$\Sigma$} & 102.67 & 287 & 402 & [98.08, 662.62] \\ 
    \textbf{Hunyuan} & 93.31 & 487 & 340 & [99.68, 416.45] \\ 
    \textbf{SDXL} & 112.44 &  803 & 447 & [101.57, 704.89] \\ 
    \textbf{SDv1.5} & 88.17 & 294 & 378 & [92.90, 584.56] \\ 
    \textbf{DiT-XL/2} & 85.02 & 201 & 356 & [64.70, 498.65] \\ 
    \bottomrule
    \end{tabular}
    }
    \caption{Denoiser search space statistics: number of sampled configurations, number of quantizable layers $\#W$ and FID range. FID is the performance 
    of the W16A16 model.} 
    \label{tab:config_table}
\end{table}

In this section we evaluate Qua$^2$SeDiMo on several T2I DMs: PixArt-$\alpha$, PixArt-$\Sigma$, Hunyuan and SDXL. Due to space constraints, additional results on SDv1.5 and DiT-XL/2 can be found in the supplementary. We apply our scheme to find cost-effective quantization configurations that minimize both FID and model size while providing some visual examples. We then compare our found quantization configurations to existing DM PTQ literature. Finally, we share some insights on the quantization sensitivity of denoiser architectures.

\subsection{Pareto Optimal Mixed-Precision Denoisers}
\label{sec:dataset_stats}

\begin{figure*}[t]
    \centering
    \includegraphics[width=6.85in]{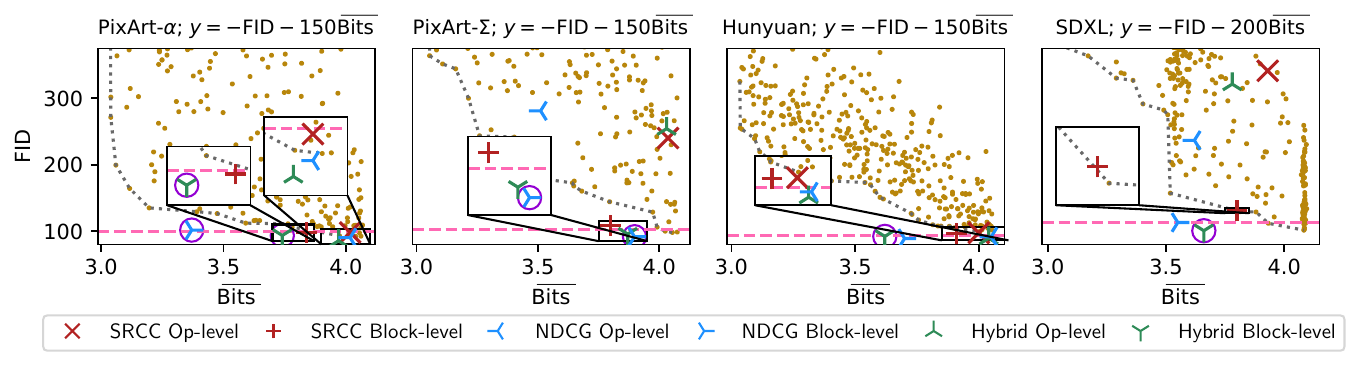}
    \caption{Results on PixArt-$\alpha$, PixArt-$\Sigma$, Hunyuan and SDXL under constrained optimization to minimize FID and $\widebar{Bits}$. Dashed horizonal line denotes the FID of the W16A16 model. Dotted grey line denotes the Pareto frontier constructed from our corpus of randomly sampled configurations (yellow dots). For each predictor ensemble, we generate two quantization configurations: `Op-level' for individual weight layers and `Block-level' for subgraph structures. Purple circles denote configurations we later investigate to generate images and draw insights from. Best viewed in color.}
    \label{fig:pareto_dit_sdxl}
\end{figure*}

To train Qua$^2$SeDiMo predictors, we sample and evaluate hundreds of randomly selected quantization configurations per denoiser architecture. To evaluate a configuration, we generate 1000 images and compute the FID score relative to a ground-truth image set. Specifically, for all T2I DMs, we use prompts and images from the COCO 2017 validation set to generate images and compute FID, respectively. For DiT-XL/2, we generate one image per ImageNet class and measure FID against the ImageNet validation set. We generate $1024^2$ images using PixArt-$\Sigma$ and Hunyuan and set a resolution of $512^2$ for all other DMs. We report additional details, e.g., number of steps, in the supplementary. Table~\ref{tab:config_table} lists statistics for each denoiser search space.

We focus on maximizing visual quality while minimizing the average bit precision $\widebar{Bits}$ of the denoiser. To achieve this, we train Qua$^2$SeDiMo to predict $y=-FID-\lambda\widebar{Bits}$. Specifically, $\lambda$ re-scales $\widebar{Bits}$ to determine how we weigh model size against performance (FID). As such, $\lambda$ is a denoiser-dependent coefficient. Further, we consider three ranking losses $\mathcal{L}_{rank}$: the differentiable spearman $\rho$ from \citet{blondel2020fast} that maximizes SRCC, LambdaRank which maximizes NDCG and a `Hybrid' loss that sums both of them to maximize SRCC and NDCG.

To leverage our limited training data per Table~\ref{tab:config_table}, we follow \citet{mills2024autobuild} and train a predictor ensemble to generate subgraph scores using different data splits. Specifically, we split the corpus of quantization configurations into $K=5$ folds, each containing an 80\%/20\% training/validation data split with disjoint validation partitions. We measure validation set performance for each predictor in the ensemble and use it as a weight to re-scale predictor scores. Detailed predictor hyperparameters and other details can be found in the supplementary. 

Finally, we construct two quantization configurations: Operation and Block-level. Operation-level optimization enumerates each weight layer nodes $v$ and selects the quantization method and bit precision that produces the highest score $\norm{h_{v_i}^0}_1$. Block-level optimization enumerates settings for all nodes in block subgraphs to maximize the score of the subgraph root node.

Figure~\ref{fig:pareto_dit_sdxl} reports our findings on PixArt-$\alpha$, PixArt-$\Sigma$, Hunyuan and SDXL for $y=-FID-\lambda\widebar{Bits}$. Additional results for for $y=-FID$ can be found in the supplementary. We observe that quantization configurations generated using the subgraph `Block-level' approach with the NDCG and Hybrid losses are consistently superior to those found using the baseline SRCC loss and the Pareto frontier of randomly sampled training configurations. Generally, `Op-level' optimization fails outright or fixates on the low-FID, high $\widebar{Bits}$ region in the bottom right corner, but in either case, fails to produce configurations that optimize $y=-FID-\lambda\widebar{Bits}$.

In terms of specific quantization configurations, on PixArt-$\alpha$, we are able to find a remarkable quantization configuration that achieves 3.4-bit precision with comparable FID to the W16A16 model. Impressively, we also find 3.7-bit configurations that outperform the W16A16 model FID on PixArt-$\alpha$ and SDXL as well as a 3.65-bit Hunyuan configuration. Finally, PixArt-$\Sigma$ proves to be the hardest denoiser to optimize as FID of random configurations rises sharply when quantizing below 4-bits, yet Qua$^2$SeDiMo is still able to construct several low-FID, 3.9-bit quantization configurations. Next, we compare our mixed-precision configurations to several prior 4-bit methods.

\subsection{Comparison with Related Literature}

We quantitatively and qualitatively compare Qua$^2$SeDiMo to several existing DM PTQ methods: Q-Diffusion, TFMQ-DM and ViDiT-Q. Specifically, we quantize weights down to 4-bits (W4) or lower, while considering three activation precision levels: A16, A8 and A6. Q-Diffusion and TFMQ-DM compute activation scales using a calibration set, while ViDiT-Q and Qua$^2$SeDiMo employ the online, patch-based technique from Microsoft's ZeroQuant~\cite{yao2022zeroquant}.

For each method, we sample 10k unique $(\texttt{caption}, \texttt{image})$ pairs from the COCO 2014 validation set and generate one image per caption and compute FID using the selected validation set images. We also compute the CLIP score~\cite{hessel2021clipscore} using the ViT-B/32 backbone and COCO validation captions. 

Table~\ref{tab:alpha_fid_clip} reports our findings on PixArt-$\alpha$. We note that how that at every activation bit precision level, the W4 configuration built by Qua$^2$SeDiMo achieves the best FID and CLIP metrics while the W3.7 and W3.4 variants are not far behind, especially in terms of CLIP score. The most competitive method is ViDiT-Q, followed by TFMQ-DM. In contrast, we deliberately re-ran Q-Diffusion using the online ZeroQuant activation quantization (Q-Diffusion OAQ) as its original mechanism catastrophically fails in the W4A8 and W4A6 settings for DiTs. 

\begin{figure*}[t!]
    \centering
    \includegraphics[width=6.9in]{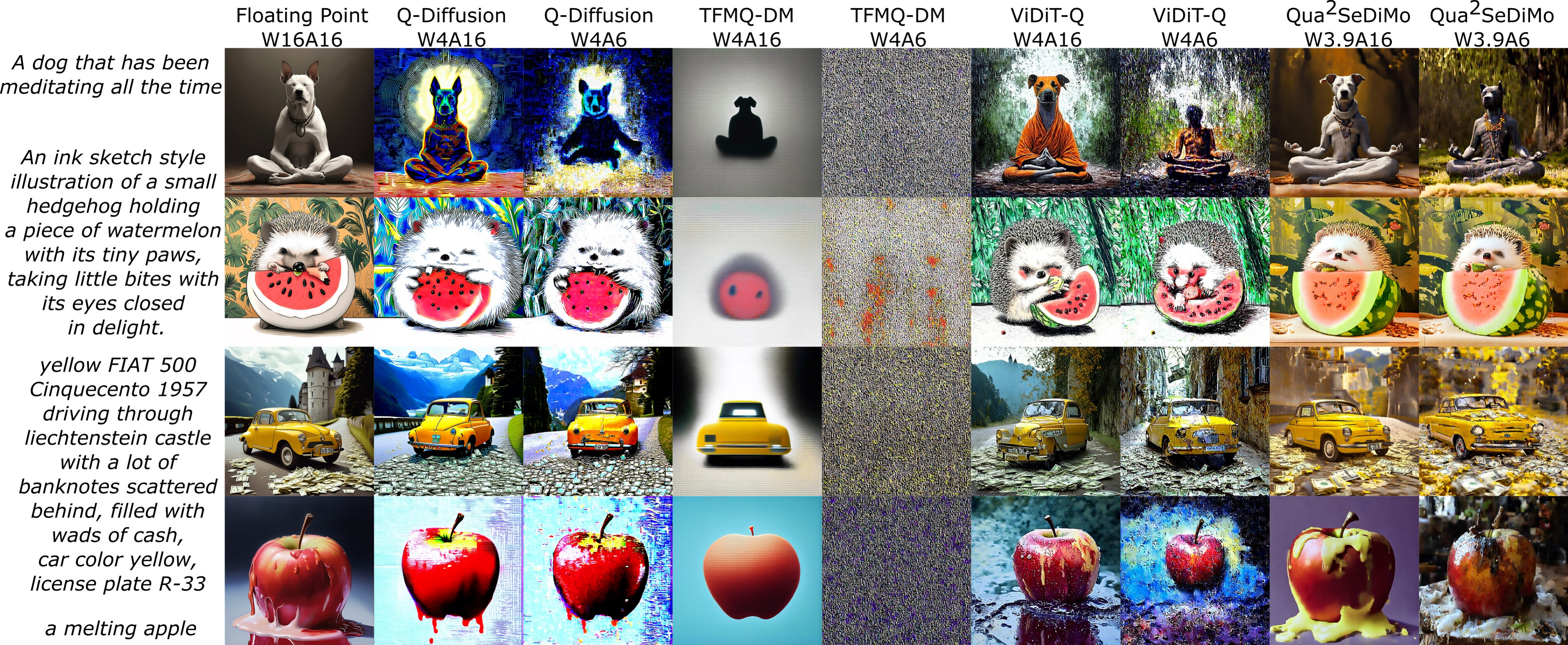}
    \caption{PixArt-$\Sigma$ example images and comparison with related work. Resolution: $1024^2$.}
    \label{fig:sigma_images}
\end{figure*}

Next, Table~\ref{tab:sigma_fid_clip} provides an analogous comparison for PixArt-$\Sigma$. This denoiser is harder to quantize than its predecessor, yet despite this we are still able to find a W3.9-bit precision quantization configuration that outperforms competiting methods across all activation precision levels. Curiously, more traditional PTQ approaches for U-Nets like Q-Diffusion are more competitive at this level for weight quantization, but must still discard calibration-based activation quantization in favour of the online approach.

\begin{table}[t!]
    \centering
    \scalebox{0.9}{
    \begin{tabular}{lccc} \toprule
    \textbf{Method} & \textbf{Precision} & \textbf{FID $\downarrow$} & \textbf{CLIP $\uparrow$} \\ \midrule
    Full Precision & W16A16 & 34.05 & 0.3102  \\ \midrule
    Q-Diffusion & W4A16 & 41.93 & \textit{0.2992} \\ 
    TFMQ-DM & W4A16 & 38.67 & 0.2905 \\
    ViDiT-Q & W4A16 & 33.98 & 0.2924 \\
    Qua$^2$SeDiMo & W4A16 & \textbf{29.02} & \textbf{0.3056} \\
    Qua$^2$SeDiMo & W3.7A16 & \textit{33.06} & 0.2931 \\
    Qua$^2$SeDiMo & W3.4A16 & 35.51 & 0.2919 \\ \midrule
    Q-Diffusion & W4A8 & 373.62 & 0.2126 \\
    Q-Diffusion OAQ & W4A8 & 52.20 & 0.2924 \\
    TFMQ-DM & W4A8 & 64.73 & 0.2594 \\
    ViDiT-Q & W4A8 & \textit{39.11} & 0.2925 \\
    Qua$^2$SeDiMo & W4A8 & \textbf{38.39} & \textbf{0.3015} \\
    Qua$^2$SeDiMo & W3.7A8 & 51.48 & \textit{0.2928} \\
    Qua$^2$SeDiMo & W3.4A8 & 53.61 & 0.2921 \\ \midrule
    Q-Diffusion & W4A6 & 382.59 & 0.2121 \\ 
    Q-Diffusion OAQ & W4A8 & 70.96 & 0.2865 \\
    TFMQ-DM & W4A6 & 90.90 & 0.2545 \\
    ViDiT-Q & W4A6 & \textit{56.54} & 0.2852 \\
    Qua$^2$SeDiMo & W4A6 & \textbf{54.59} & \textbf{0.2950} \\
    Qua$^2$SeDiMo & W3.7A6 & 58.05 & \textit{0.2906} \\
    Qua$^2$SeDiMo & W3.4A6 & 59.75 & 0.2898 \\ \bottomrule
    \end{tabular}
    }
    \caption{Quantization comparison on PixArt-$\alpha$ generating 10k $512^2$ images using COCO 2014 prompts. Q-Diffusion OAQ pairs the original method with online activation quantization. Best/second best results in bold/italics. }
    \label{tab:alpha_fid_clip}
\end{table}

\begin{table}[t!]
    \centering
    \scalebox{0.9}{
    \begin{tabular}{lccc} \toprule
    \textbf{Method} & \textbf{Precision} & \textbf{FID $\downarrow$} & \textbf{CLIP $\uparrow$} \\ \midrule
    Full Precision & W16A16 & 36.94 & 0.3154  \\ \midrule
    Q-Diffusion & W4A16 & 38.27 & 0.3115 \\ 
    TFMQ-DM & W4A16 & 39.15 & 0.3096 \\
    ViDiT-Q & W4A16 & \textit{31.21} & \textit{0.3132} \\
    Qua$^2$SeDiMo & W3.9A16 & \textbf{30.34} & \textbf{0.3154} \\ \midrule
    Q-Diffusion & W4A8 & 533.49 & 0.2219 \\ 
    Q-Diffusion OAQ & W4A8 & \textit{36.89} & \textit{0.3127} \\
    TFMQ-DM & W4A8 & 62.98 & 0.2987 \\
    ViDiT-Q & W4A8 & 37.17 & 0.3072 \\
    Qua$^2$SeDiMo & W3.9A8 & \textbf{35.61} & \textbf{0.3136} \\ \midrule
    Q-Diffusion & W4A6 & 533.49 & 0.2117 \\
    Q-Diffusion OAQ & W4A6 & \textit{74.83} & \textit{0.2981} \\ 
    TFMQ-DM & W4A6 & 154.13 & 0.2600 \\
    ViDiT-Q & W4A6 & 87.47 & 0.2837 \\
    Qua$^2$SeDiMo & W3.9A6 & \textbf{74.29} & \textbf{0.2999} \\ \bottomrule
    \end{tabular}
    }
    \caption{Quantization comparison on PixArt-$\Sigma$ generating 10k $1024^2$ images using COCO 2014 prompts. Same experimental setup as Table~\ref{tab:alpha_fid_clip}.} 
    \label{tab:sigma_fid_clip}
\end{table}

\begin{table}[t!]
    \centering
    \scalebox{0.9}{
    \begin{tabular}{l|l|c|c} \toprule
    \textbf{Model} & \textbf{Method} & \textbf{Precision} & \textbf{\#Votes} (\%) $\uparrow$ \\ \midrule 
    \multirow{5}{*}{PixArt-$\alpha$} & Q-Diffusion & W4A8 & \textit{34 (28.81\%)}  \\ 
    & TFMQ-DM & W4A8 & 10 (8.47\%) \\
    & ViDiT-Q & W4A8 &10 (8.47\%) \\
    & Qua$^2$SeDiMo & W4A8 & \textbf{51 (43.22\%)} \\
    \midrule 
    \multirow{5}{*}{PixArt-$\Sigma$} & Q-Diffusion &W4A8 & \textit{28 (23.73\%)}  \\ 
    & TFMQ-DM & W4A8 &3 (2.54\%)  \\
    & ViDiT-Q & W4A8 &\textit{28 (23.73\%)}\\
    & Qua$^2$SeDiMo & W3.9A8 &\textbf{44 (37.29\%)} \\
    \midrule
    \end{tabular}
    }
    \caption{User preference study between Qua$^2$SeDiMo and baseline methods on PixArt-$\alpha$/$\Sigma$. 
    $N=118$. Best/second best results in bold/italics.}
    \label{tab:user_preference}
\end{table}

In terms of qualitative comparison, recall Fig.~\ref{fig:alpha_images} which shows generated images on PixArt-$\alpha$, while Figure~\ref{fig:sigma_images} provides images for PixArt-$\Sigma$. We note the robustness of Qua$^2$SeDiMo, as even the sub 4-bit configurations can generate acceptable images with low-bit activation quantization.

\begin{figure*}[t!]
    \centering
    \includegraphics[width=6.71in]{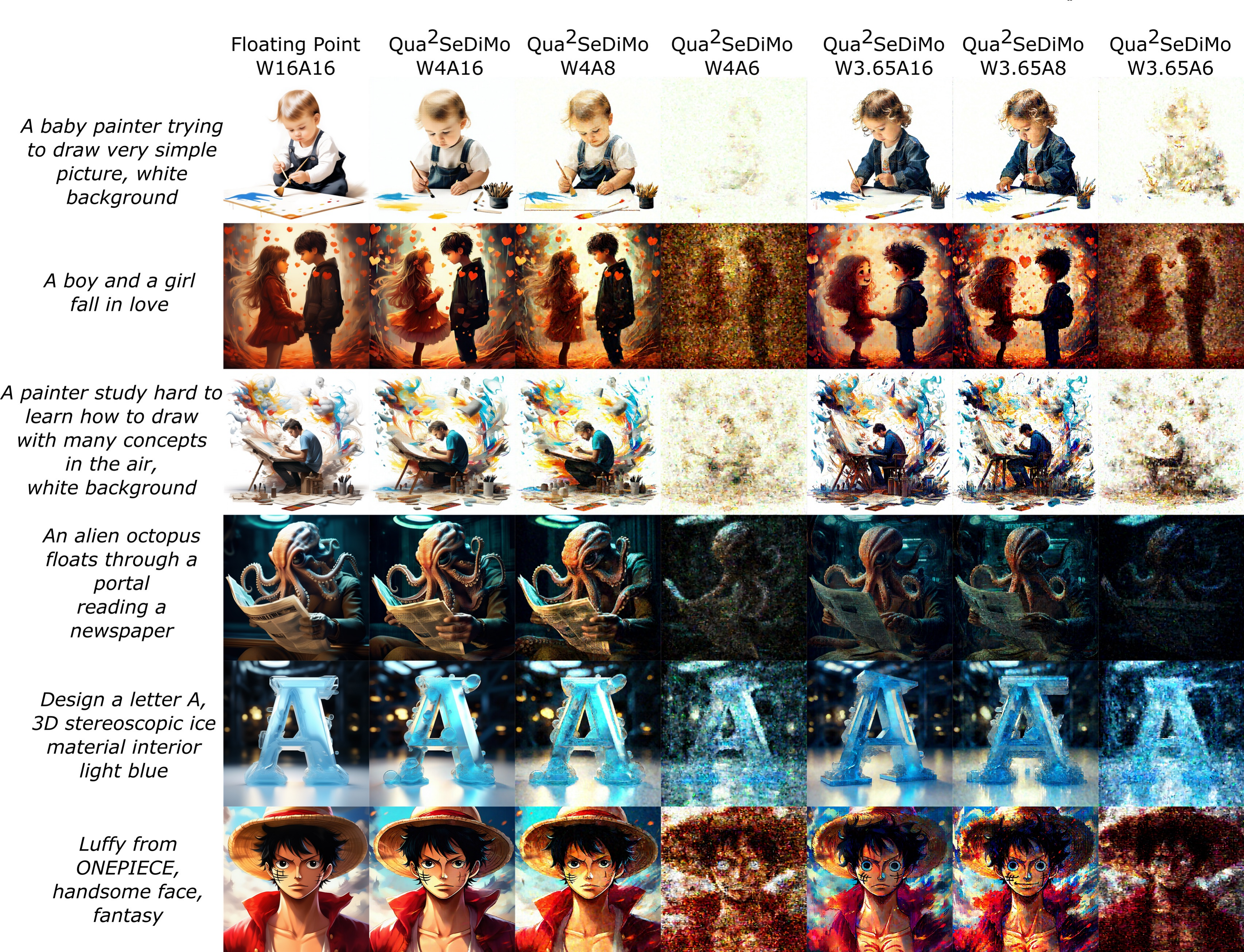}
    \caption{Hunyuan-DiT example images. Resolution: $1024^2$.} 
    \label{fig:hunyuan_images}
\end{figure*}

Finally, Table~\ref{tab:user_preference} provides the results of a human preference study qualitatively comparing images produced by Qua$^2$ SeDiMo with other methods. These studies consisted of 20 human participants and 118 images. Each participant was given a prompt and the corresponding generated images for four W4A8 models quantized by different methods and asked to choose which image was best in terms of visual quality and prompt adherence. Users were given a `Cannot Decide' option but asked to invoke it sparingly (13 times for $\alpha$ \& 15 for $\Sigma$). The results of this survey show a significant preference for the images produced by Qua$^2$SeDiMo compared to other approaches for both PixArt models.

\subsection{Results on Hunyuan-DiT}

\begin{table}[t!]
    \centering
    \scalebox{0.9}{
    \begin{tabular}{lccc} \toprule
    \textbf{Method} & \textbf{Precision} & \textbf{FID $\downarrow$} & \textbf{CLIP $\uparrow$} \\ \midrule
    Full Precision & W16A16 & 41.92 & 0.3089  \\ \midrule
    Q-Diffusion   & W4A16 & 42.09 & 0.3095 \\
    TFMQ-DM       & W4A16 & 42.50 & 0.3066 \\
    Qua$^2$SeDiMo & W4A16 & \textbf{40.25} & \textbf{0.3162} \\
    Qua$^2$SeDiMo & W3.65A16 & 41.97 & 0.3158 \\ \midrule
    Q-Diffusion OAQ & W4A8  & 71.99 & 0.2974 \\
    TFMQ-DM       & W4A8  & 72.15 & 0.2929 \\
    Qua$^2$SeDiMo & W4A8 &\textbf{ 51.91} & \textbf{0.3158} \\
    Qua$^2$SeDiMo & W3.65A8 & 71.32 & 0.3078 \\ \midrule
    Q-Diffusion OAQ & W4A6 & 171.40 & 0.2395 \\
    TFMQ-DM       & W4A6 & 175.32 & 0.2375 \\ 
    Qua$^2$SeDiMo & W4A6 & \textbf{155.19} & \textbf{0.2438} \\
    Qua$^2$SeDiMo & W3.65A6 & 167.67 & 0.2322 \\ \bottomrule
    \end{tabular}
    }
    \caption{Quantization comparison on Hunyuan-DiT generating 10k $1024^2$ images using COCO 2014 prompts. Same experimental setup as Table~\ref{tab:sigma_fid_clip}. Best result in bold.}
    \label{tab:hunyuan_fid_clip}
\end{table}

\begin{figure*}[t!]
    \centering
    \includegraphics[height=1.05in]{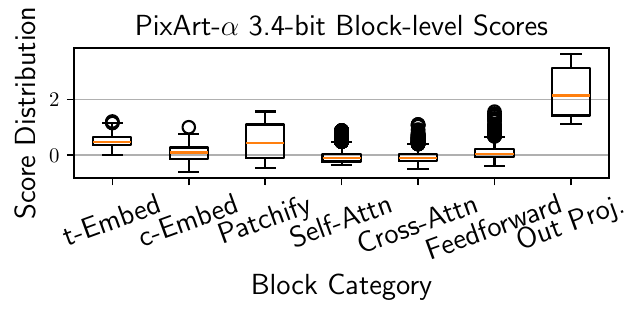}
    \includegraphics[height=1.05in]{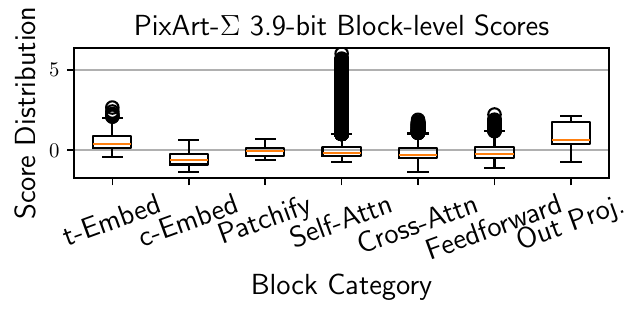}
    \includegraphics[height=1.05in]{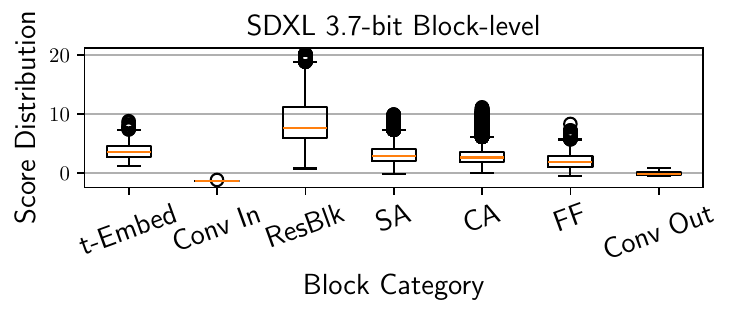}
    \caption{Block-level box-plots for sub 4-bit PixArt-$\alpha$, PixArt-$\Sigma$ 
    and SDXL configurations.}
    \label{fig:box_plots_main}
\end{figure*}

Table~\ref{tab:hunyuan_fid_clip} compares Qua$^2$SeDiMo to other methods on Hunyuan. We observe better performance in terms of lower FID and higher CLIP at W4 and W3.65-bit precision. However, compared to PixArt DiTs, Hunyuan is more difficult to adequately quantize to A6-bit precision, as all methods experience a substantial performance degradation at this level.

This degradation is visualized in Figure~\ref{fig:hunyuan_images}, which provides images generated by Hunyuan when quantized by Qua$^2$SeDiMo. Specifically, we examine images at W\{4, 3.65\}A\{16, 8, 6\}-bit precision levels. This comparison visually contrasts the effect of weight and activation quantization. Specifically, weight quantization controls higher-level aspects of an image, e.g., the child's hair and clothing, artstyle of the boy and girl, shape of the octopus' head and Luffy's facial expression. In contrast, there is an inverse relationship between the activation bit precision and the amount of undesirable noise present.

\subsection{Extracted Insights}

We examine some of the quantization sensitivity insights Qua$^2$SeDiMo provides. Figure~\ref{fig:box_plots_main} plots the sensitivity score distributions for different subgraph block types, e.g., Self-Attention (SA) or Cross-Attention (CA). We interpret these scores as follows: If the score distribution for a block type has a large range with high outliers, it means there are quantization block settings which are crucial to maintaining efficient performance. If the distribution mean and range are low, the block is not very important.

Corroborating \citet{huang2023tfmq}, we find that the time embedding module (t-Embed) is an important block as the score distribution for each denoiser has a large mean, wide range, and a number of high-scoring outliers. In the SDXL U-Net, the time parameter interfaces with each convolutional ResNet Block (ResBlk), which carries the highest score distribution for that denoiser. In contrast, the condition embedding (c-Embed) in PixArt-$\alpha$/$\Sigma$ is quite low, indicating that adequate quantization of prompt embedding layers is less crucial. Also, note the moderate variance in the input `Patchify' and `Out Proj.' layers of PixArt DMs, indicating great importance, especially in contrast to the analogous `Conv In' and `Conv Out' in SDXL.

\begin{figure}[t!]
    \centering
    \includegraphics[width=3.1in]{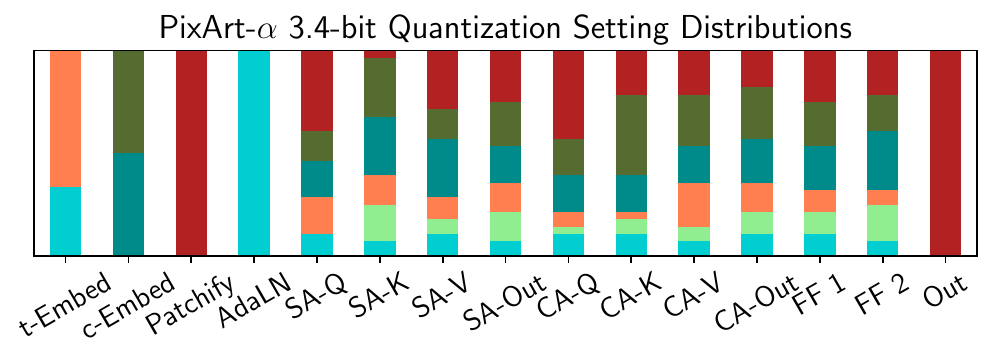}
    \includegraphics[width=3.1in]{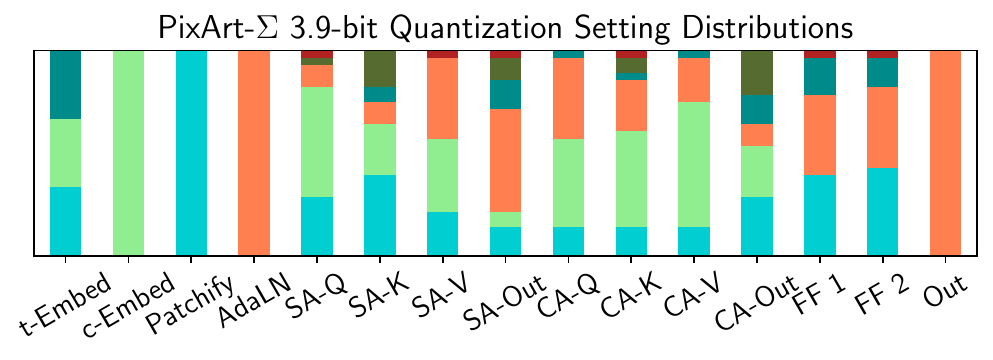}
    \includegraphics[width=3.1in]{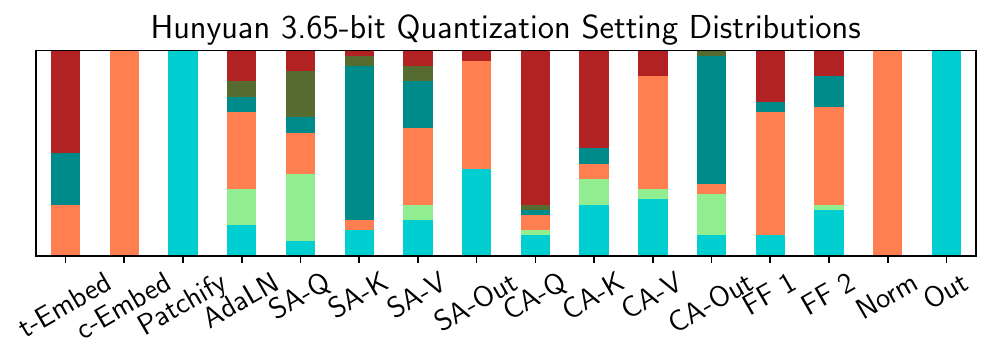}
    \hspace*{0.08in}\includegraphics[width=3.175in]{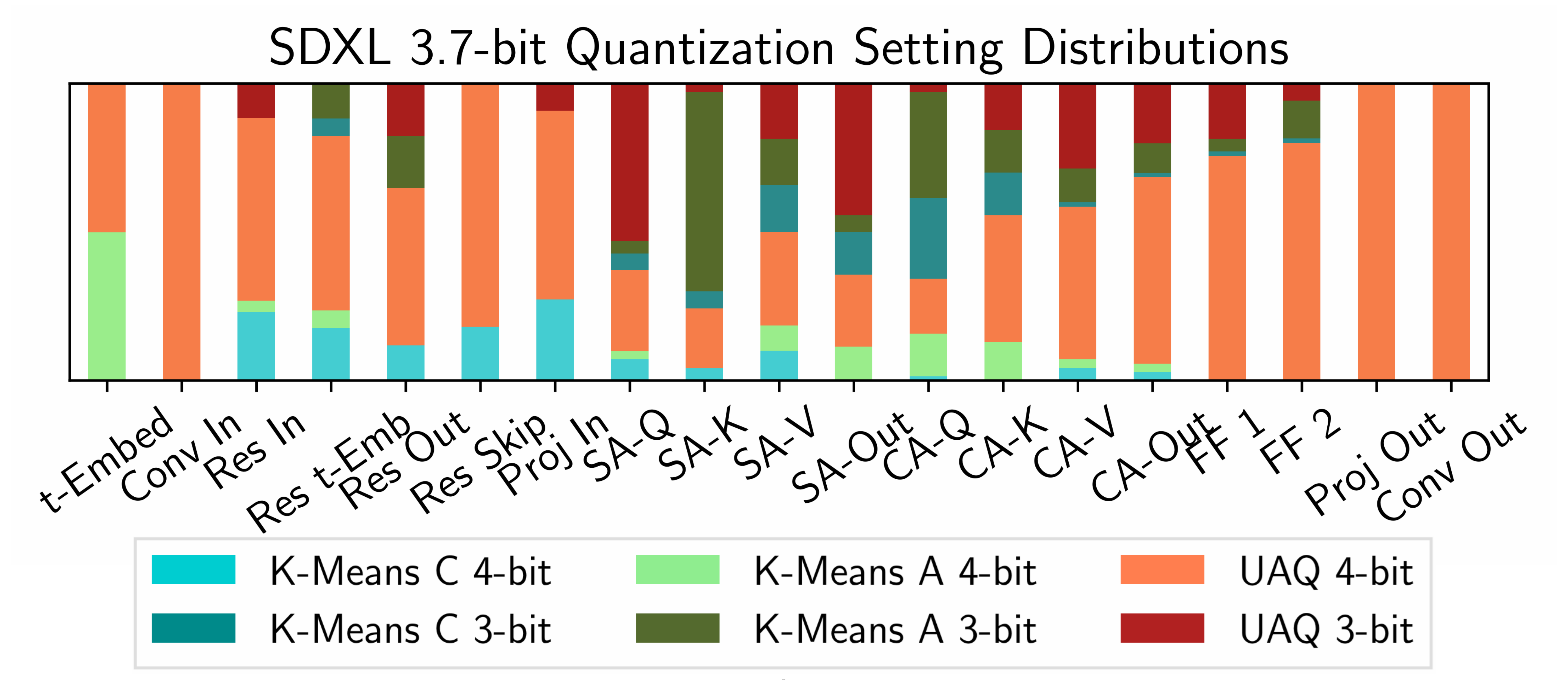} 
    \caption{
    Stacked quantization method bar plots for several sub 4-bit quantization configurations. Best viewed in color.}
    \label{fig:bar_plots_main}
\end{figure}

Finally, Figure~\ref{fig:bar_plots_main} shows stacked bar plots illustrating the distribution of quantization methods and bit precisions selected to form the optimal sub 4-bit configurations. That is, PixArt-$\alpha$ contains 4 t-Embed linear layers, all kept at 4-bit precision: 3 using UAQ, and one using $K$-Means C. The model also contains 28 self-attention key (SA-K; one for each transformer block) layers quantized primarily using $K$-Means C/A at 3 and 4-bit precisions. It also has a single output (Out) layer quantized to 3-bits using UAQ. In general, these findings show that DiT blocks have a slight preference for $K$-Means-based quantization, whereas by contrast, the SDXL U-Net strongly prefers UAQ quantization. 

%% file: src/conclusion.tex
\section{Conclusion}
\label{sec:conclusion}

We propose Qua$^2$SeDiMo, a mixed-precision DM weight PTQ framework. We cast denoisers as large search spaces characterized by choice of bit precision and quantization method per weight layer. It extracts quantifiable insights about how these choices correlate to end-to-end metrics such as FID and average bit precision. We use these insights to construct high-quality sub 4-bit weight quantization configurations for several popular T2I denoisers such as PixArt-$\alpha$/$\Sigma$, Hunyuan and SDXL. We pair this method with low-bit activation quantization to outperform existing methods and generate convincing visual content. 

%% file: src/appendix.tex
\section{Supplementary Appendix}

First, we provide details of the denoiser subgraphs used for each DM. Then we provide a slew of additional experimental results on SDXL, SDv1.5 and DiT-XL/2, including sampled visual results. We also provide numerous insights and charts quantifying the quantization sensitivity of each DM. Finally, we provide additional experimental details, hyperparameter settings and detail our compute setup.

\subsection{Denoiser Inference Hyperparameters}

We build our code base on top of the open-source repository provided by Q-Diffusion, which provides support for quantization and inference on SDv1.5. SDv1.5 denoises for 50 timesteps with a Classifier-Free Guidance (CFG) scale of 7.5 by default. For all other Diffusion Models we rely on the open-source implementation and default hyperparameters provided by HuggingFace Diffusers~\cite{von-platen-etal-2022-diffusers}: PixArt-$\alpha$ and PixArt-$\Sigma$ denoise for 20 steps each using a CFG scale of 4.5. Hunyuan-DiT denoises over 50 steps with a default CFG scale of 5.0. For SDXL we utilize both the base and refiner U-Nets: Inference takes 40 steps, split 32/8 between the base and refiner, respectively, with a guidance scale of 5.0. Note that we only quantize the SDXL base U-Net. DiT images denoise over 250 timesteps with a default CFG scale of 4.0. Finally, We generate $1024^2$ images using PixArt-$\Sigma$ and Hunyuan-DiT and $512^2$ images using all other DMs.

\subsection{Additional Quantization Details}
We now enumerate some additional experimental details related to our quantization implementation, quantifying the floating point overhead of different methods, and how to sample mixed-precision, mixed method quantization configurations from a denoiser search space.
We build our code base on top of Q-Diffusion which applies simulated quantization to each individual weight layer in a DM denoiser architecture by rounding and binning floating point weights and activations. We extend this code base to provide support for $K$-Means clustering quantization. We also extend the implementation to support denoisers beyond SDv1.4/1.5, including extending the channel-splitting approach for quantizing U-Net long residual connections to support Hunyuan-DiT and SDXL as well as SDv1.5. 

\subsubsection{Quantifying Quantization Floating Point Overhead. }
\label{app:overhead}

We provide a formal calculation of the floating point overhead imposed by the $K$-Means and UAQ quantization techniques. For $K$-Means, each element in $W_Q$ is an $N_Q$-bit index corresponding to one of $2^{N_Q}$ cluster centroids, each with precision $N_{FP}$. Therefore, the total bits $b_K$ required to store and dequantize a compressed weight tensor is 

\begin{equation}
    \centering
    \label{eq:k_means}
    b_{K} = \texttt{size}(W_{FP})N_Q + N_{FP}2^{N_{Q}}\sigma_K,
\end{equation}
where \texttt{size} returns the number of elements in a tensor and $\sigma_K$ can be either $c_{out}$ or 1 depending on if quantization is performed channel-wise or across the entire tensor, respectively. 

For UAQ, we only need to store the scale $\Delta$ potentially an asymmetric zero-point $z$ in $N_{FP}$ precision. This form of quantization is done once per output channel $c_{out}$, so the number of bits $b_{UAQ}$ is given by 

\begin{equation}
    \centering
    \label{eq:uaq_b}
    b_{UAQ} = \texttt{size}(W_{FP})N_Q + N_{FP}c_{out}\sigma_{z},
\end{equation}
where $\sigma_z = 2$ is 2 if an asymmetric zero-point is used and $1$ otherwise.

\subsubsection{Single Quantization Method Application. }

\begin{table*}[t!]
    \centering
    \scalebox{0.9}{
    \begin{tabular}{lccccccc} \toprule
    \textbf{Quant. Method} & \textbf{Full Precision} & & \textbf{W4A16} &  & & \textbf{W3A16} & \\ \midrule
    \textbf{Denoiser} & \textbf{W16A16} & \textbf{K-Means C} & \textbf{K-Means A} & \textbf{UAQ} & \textbf{K-Means C} & \textbf{K-Means A} & \textbf{UAQ} \\ \midrule
    \textbf{PixArt-$\alpha$} & 99.67 & 97.88 & 353.96 & 97.45 & 178.59 & 430.11 & 172.51 \\
    \textbf{PixArt-$\Sigma$} & 101.67 & 103.11 & 346.65 & 230.77 & 219.38 & 414.53 & 599.78 \\
    \textbf{Hunyuan} & 93.31 & 96.93 & 186.44 & 108.03 & 130.96 & 358.15 & 182.87 \\
    \textbf{SDXL} & 112.44 & 120.43 & 243.51 & 309.25 & 423.85 & 423.85 & 463.74 \\
    \textbf{DiT-XL/2} & 85.02 & 77.79 & 420.45 & 67.48 & 109.06 & 429.28 & 448.92 \\
    \textbf{SDv1.5} & 88.17 & 86.09 & 155.76 & 410.48 & 369.49 & 355.87 & 434.79 \\ \bottomrule
    \end{tabular}
    }
    \caption{FID scores for all DM denoiser architectures considered in this paper when all weight layers are uniformly quantized to one method and bit precision, e.g., 3-bit UAQ. For each quantization configuration, we generate 1k images either using MS-COCO prompts (for all denoisers except DiT-XL/2) or one image per ImageNet class (DiT-XL/2).}
    \label{tab:uniform}
\end{table*}

We consider two bit-precision levels $\{3, 4\}$ and three quantization methods, $K$-Means C, $K$-Means A and UAQ in this paper for a total of six options per weight layer. We generate search spaces from these options and use them to find effective mixed-precision weight quantization configurations. However, one might be interested to know what kind of performance is obtainable when applying each of these options \textit{uniformly} across every weight layer in the DM denoiser network. Table~\ref{tab:uniform} reports these findings for each denoiser considered in this paper. Note how at 4-bit precision, the most effective method is generally $K$-Means C. While UAQ can sometimes outperform $K$-Means C, e.g., on DiT-XL/2, there are times when it severely underperforms, e.g., on PixArt-$\Sigma$ and SDXL. $K$-Means A is ineffective than either, but as Figures~\ref{fig:bar_plots_main} and \ref{fig:bar_plots_app} show, it can help produce very efficient quantization configurations when reserved for certain, niche weight layer types. 

\subsubsection{Sampling Quantization Configurations. }
\label{app:sampling}
To sample a quantization configuration, we first draw a value $p \in [0, 1]$ from a uniform distribution. $p$ is the Bernoulli probability that a given weight layer will be quantized to 3-bits, i.e., if $p=1$, the entire quantization configuration will be set to 3-bits. We then enumerate each quantizable weight-layer in the denoiser, using $p$ to determine the bit precision. We then randomly select the quantization method for each weight layer.  

\subsubsection{Encoding Quantization Configurations as Directed Acyclic Graphs. }
\label{app:encoding}
Each search space has a separate, fixed DAG structure. All variation in sampled quantization configuration stem from node features. Specifically, 
we encode the bit precision, quantization method, quantization error $\epsilon$, and size ratio $\sfrac{\texttt{size}(W_Q)}{\texttt{size}(W_{FP})}$ as node features for every search space. We also encode denoiser architecture-specific features such as weight \textit{type}, e.g., the input conv, output conv, or time-step embedder `t-Embed' of a U-Net ResNet block and position information, e.g., which ResNet block is the weight a layer part of. We encode the quantization error $\epsilon$ and size ratio as scalar values, while all other features are categorical.

\begin{figure*}[t!]
    \includegraphics[width=6.85in]{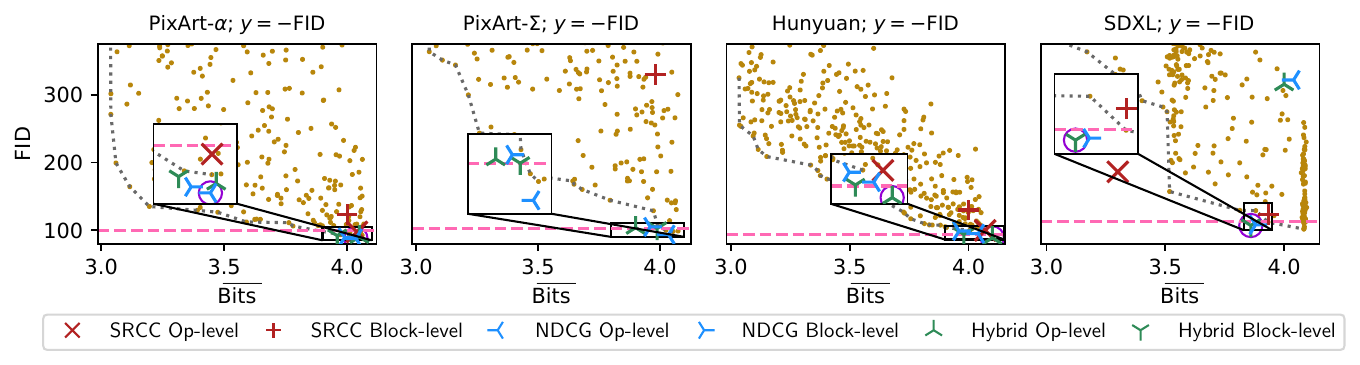}
    \caption{Results on PixArt-$\alpha$, PixArt-$\Sigma$, Hunyuan and SDXL maximing $y=-FID$ for pure performance. Dashed horizonal line denotes the FID of the W16A16 model. Dotted grey line denotes the Pareto frontier constructed from our corpus of randomly sampled configurations (yellow dots). For each predictor ensemble, we generate two quantization configurations: `Op-level' for individual weight layers and `Block-level' for subgraph structures. Purple circles denote configurations we later investigate to generate images and draw insights from. Best viewed in color.}
    \label{fig:pareto_alpha_sigma_hunyuan_xl_fid}
\end{figure*}

\begin{figure*}[t!]
    \includegraphics[width=6.85in]{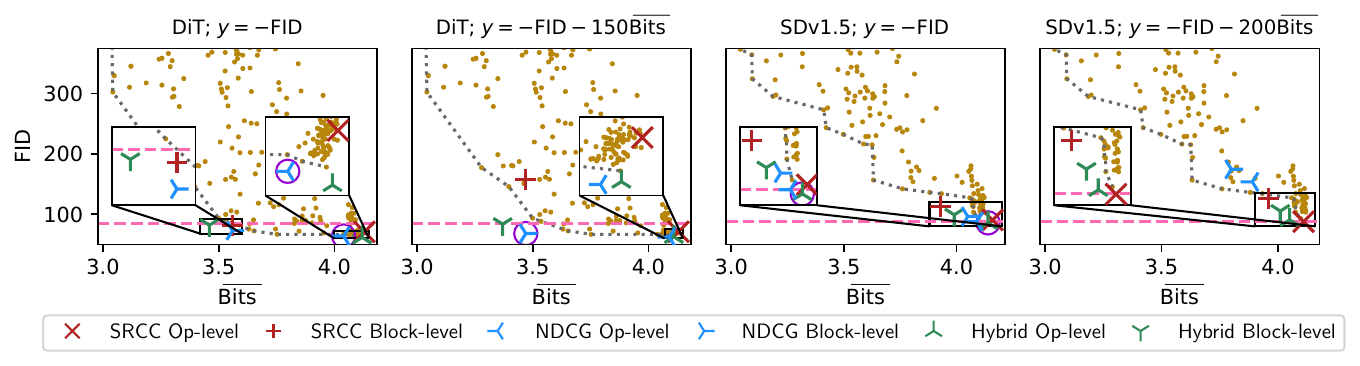}
    \caption{Results on DiT-XL/2 and SDv1.5 for pure performance $y=-FID$ and under constrained optimization to minimize FID and $\widebar{Bits}$. Same setup as Figures~\ref{fig:pareto_dit_sdxl} and \ref{fig:pareto_alpha_sigma_hunyuan_xl_fid}. Best viewed in color.}
    \label{fig:pareto_dit_v15}
\end{figure*}

\subsection{Additional Pareto Frontiers and SDv1.5 Results}

Figure~\ref{fig:pareto_alpha_sigma_hunyuan_xl_fid} provides FID-$\widebar{Bits}$ Pareto frontiers for PixArt-$\alpha$, PixArt-$\Sigma$, Hunyuan and SDXL for Qua$^2$SeDiMo when we optimize for pure FID score, e.g., $y=-FID$ (instead of $y=-FID-\lambda\widebar{Bits}$ per Fig.~\ref{fig:pareto_dit_sdxl}). Note how the best results are generally found either using LambdaRank `NDCG' or the Hybrid loss for $\mathcal{L}_{rank}$ instead of the SRCC loss. Also, unlike the constrained experiments in Fig.~\ref{fig:pareto_dit_sdxl}, generally both the `Op-level' and `Block-level' optimizations find similar quantization configurations in terms of performance and model size. 

Also, note that the PixArt-$\alpha$ `NDCG Op-level' result circled in purple is the 4-bit result used in Figure~\ref{fig:alpha_images}, Table~\ref{tab:alpha_fid_clip} and elsewhere. The same is true for the found quantization configurations for Hunyuan and SDXL that are circled in purple. 

Further, Figure~\ref{fig:pareto_dit_v15} illustrates our Pareto frontier results for DiT-XL/2 and SDv1.5 in both the unconstrained $y=-FID$ and constrained optimization $y=-FID-\lambda\widebar{Bits}$ scenarios. Note that DiT-XL/2 is the only non-T2I model we consider in this paper. Rather, instead of using prompts, it is class-conditional for ImageNet~\cite{deng2009imagenet} and coincidentally it is the easiest to quantize as there are many randomly sampled quantization configurations that achieve lower FID than the W16A16 baseline that have less than 3.75-bits on average. As a result we are able to find 3.5-bit weight quantization configurations using `Block-level' optimization when $y=-FID$, as well as two low-FID quantization configurations with fewer than 3.5-bits when $y=-FID-\lambda\widebar{Bits}$. 

In contrast, SDv1.5 is one of the hardest DMs to quantize below 4-bits on average. Like PixArt-$\Sigma$, the FID of randomly sampled quantization configurations sharply rises as the average number of bits drops below 4.0, causing the FID to quickly exceed that of the full precision baseline. Nevertheless, we are still able to find many low-FID 4-bit quantization configurations. 

Finally, Table~\ref{tab:sdv15_fid_clip} provides results for SDv1.5, comparing the 4-bit quantization configuration build by Qua$^2$SeDiMo to Q-Diffusion and TFMQ-DM. Once again, we note that our method achieves better FID and CLIP performance. Also, note how the FID of full precision W16A16 SDv1.5 model is substantially lower than that of PixArt-$\alpha$ (Tab.~\ref{tab:alpha_fid_clip} despite using the same set of 10k prompts. The finding demonstrates the sensitivity of the FID metric itself to the choice of prompts and base model in addition to number of $(\texttt{caption}, \texttt{image})$ pairs. 

\begin{table}[t!]
    \centering
    \scalebox{0.9}{
    \begin{tabular}{lccc} \toprule
    \textbf{Method} & \textbf{Precision} & \textbf{FID $\downarrow$} & \textbf{CLIP $\uparrow$} \\ \midrule
    Full Precision & W16A16 & 18.32 & 0.3150  \\ \midrule
    Q-Diffusion   & W4A16 & 18.71 & 0.3143 \\
    TFMQ-DM       & W4A16 & 18.38 & 0.3146 \\
    Qua$^2$SeDiMo & W4A16 & \textbf{18.28} & \textbf{0.3148} \\ \bottomrule
    \end{tabular}
    }
    \caption{Quantization comparison for SDv1.5 generating 10k $512^2$ images using COCO 2014 prompts. Same experimental setup as Tables~\ref{tab:alpha_fid_clip}/\ref{tab:sigma_fid_clip}. Specifically, we compare the FID and CLIP of a W4 quantization configuration found by Qua$^2$SeDiMo at 16-bit precision levels to that of the full precision baseline. Best result in bold.} 
    \label{tab:sdv15_fid_clip}
\end{table}

\subsection{Additional Quantization Sensitivity Insights}

\begin{figure}[t!]
    \centering
    \includegraphics[height=1.165in]{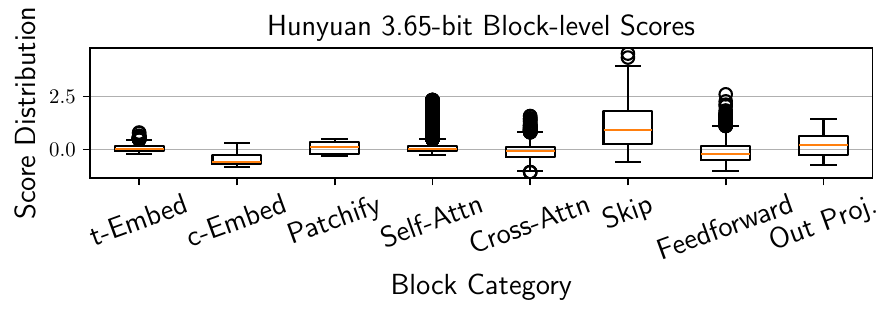}
    \caption{Block-level box-plots for the 3.65-bit Hunyuan-DiT configuration.}
    \label{fig:hunyuan_graph_op_box_365}
\end{figure}

We now provide a slew of additional quantization insight figures. First, Figure~\ref{fig:hunyuan_graph_op_box_365} extends Figure~\ref{fig:box_plots_main} for the 3.65-bit Hunyuan quantization configuration built by Qua$^2$SeDiMo. Note the large range and number of outliers associated with the `Skip' connection weight layers, highlighting that the `bimodal activation distribution' identified by Q-Diffusion~\cite{li2023q} is not limited to U-Nets. Also, similar to PixArt-$\alpha$ and PixArt-$\Sigma$, note the high median and variance associated with the `Patchify' and `Out Proj.' blocks. Finally, we compare the score distribution for time-step `t-Embed' and caption `c-Embed' embeddings, and note the importance of the former.

\subsubsection{Diffusion Transformer Op/Block-wise Sensitivity. } 
\begin{figure*}[t!]
    \centering
    \includegraphics[height=1.05in]{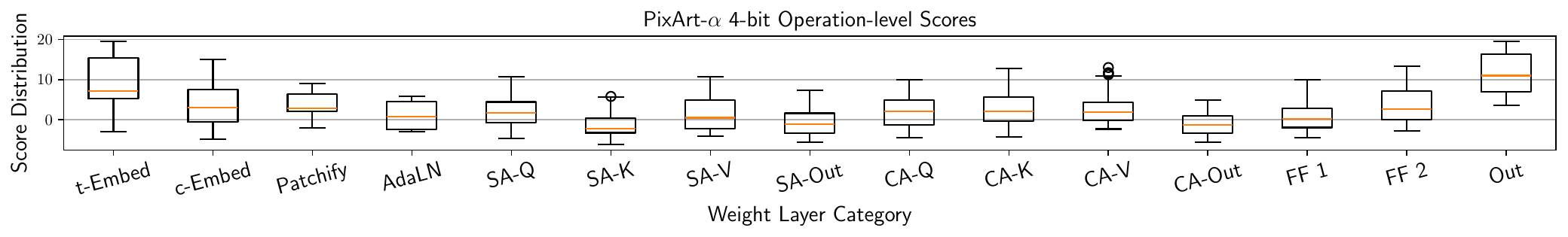}
    \includegraphics[height=1.05in]{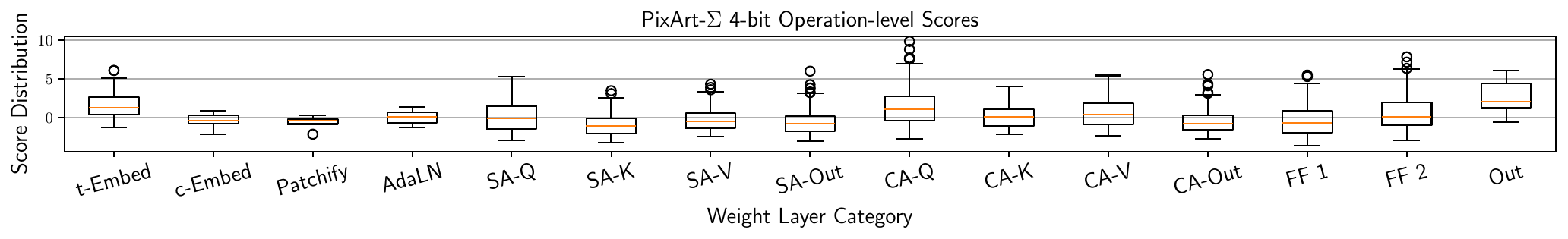}
    \includegraphics[height=1.05in]{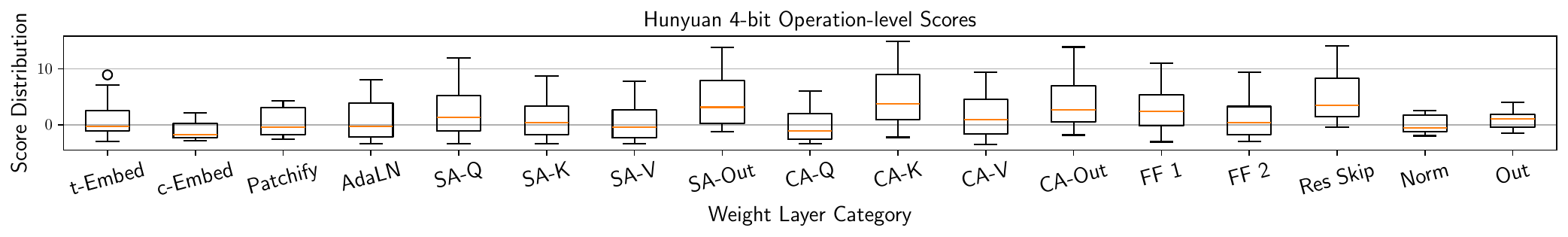}
    \includegraphics[height=1.05in]{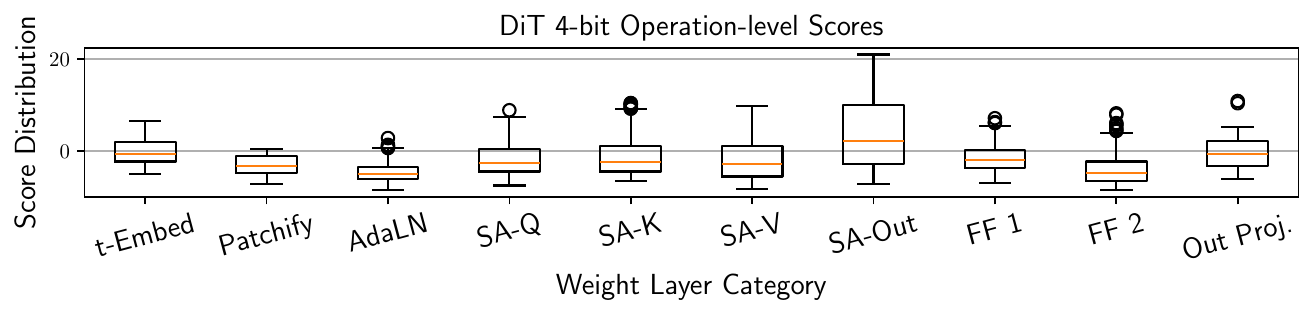}
    \includegraphics[height=1.05in]{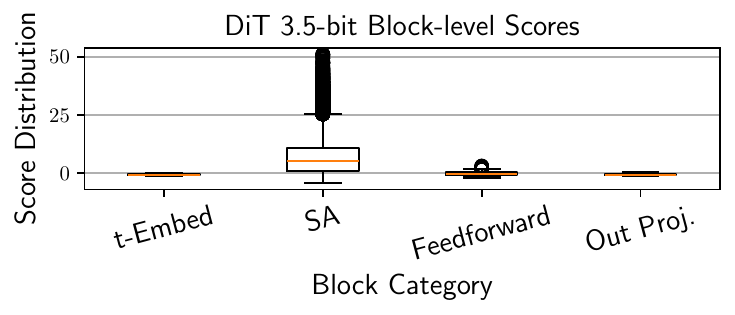}
    \caption{Operation-wise score box plots for several DiT models at multiple weight bit-precision levels. Note several acronyms: `SA', `CA' and `FF' mean `Self-Attention', `Cross-Attention' and `Feedforward', respectively.}
    \label{fig:dit_opwise}
\end{figure*}

We provide additional operation-wise and block-wise sensitivity results in Figure~\ref{fig:dit_opwise}. Note the uneven distribution of importance sensitivity scores amongst the different parts of the Transformer block. One interesting and consistent finding is the importance of the initial time-step embedding `t-Embed' layers compared to the Adaptive LayerNorm (AdaLN)~\cite{perez2018film} it directly feeds into. Generally, `t-Embed' is one of the most important non-attention weight layers alongside the final output projection `Out Proj.' for all four denoisers. By contrast, the score distributions for AdaLN is lower in terms of median value and range, indicating a lesser importance. 

The one exception is DiT-XL/2 at 4-bit precision, which has several high-velue outlier scores (represented as circles). This is likely because it handles the class-conditional embedding in additional to time-steps, whereas the PixArt and Hunyuan DiT architectures employ cross-attention mechanisms. Additionally, there is no real consistent pattern amongst the four types of DiTs as to which specific attention and feedforward weight layers are more important: While `FF 2' has higher scores than `FF 1' for PixArt-$\alpha$ and PixArt-$\Sigma$, the reverse is true for Hunyuan-DiT and DiT-XL/2, on average. Finally, the weights corresponding to long skip connections in Hunyuan, `Res Skip' possess a high score distribution and median, reflecting their relevance.

\begin{figure*}[t!]
    \centering
    \includegraphics[height=1.05in]{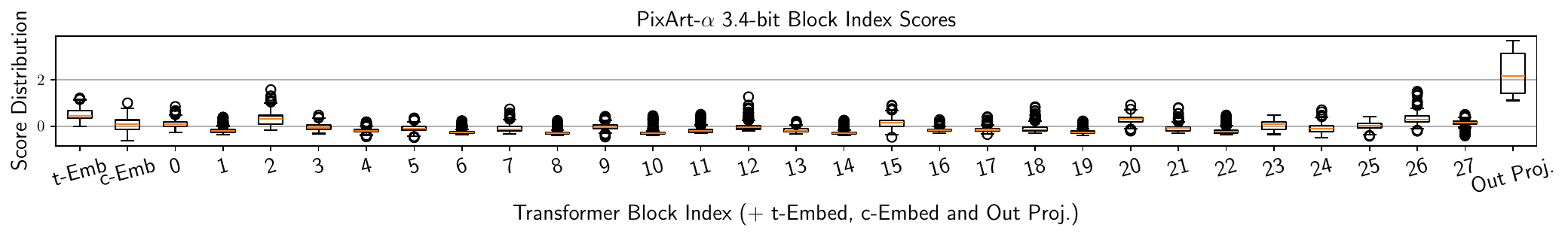}
    \includegraphics[height=1.05in]{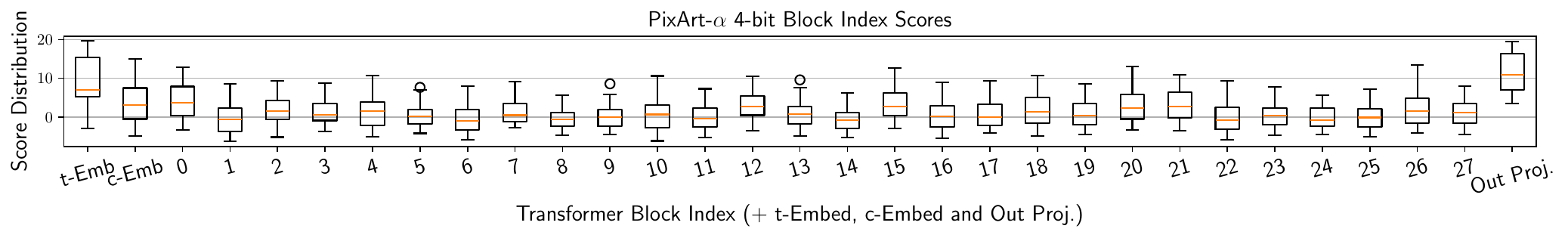}
    \includegraphics[height=1.05in]{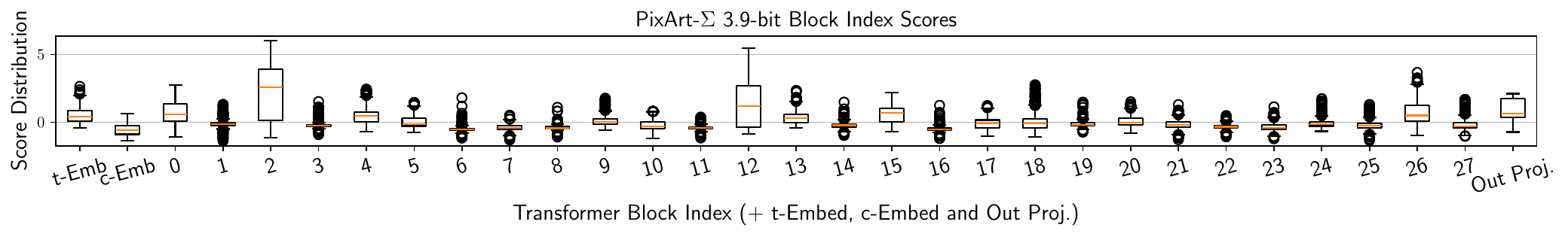}
    \includegraphics[height=1.05in]{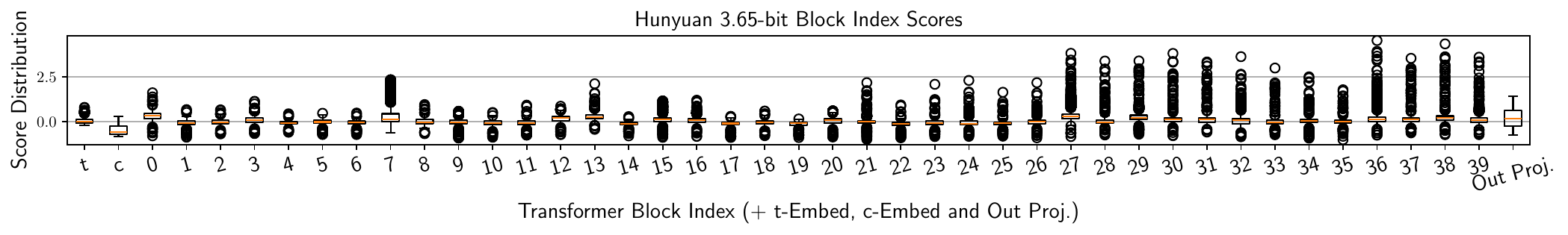}
    \includegraphics[height=1.05in]{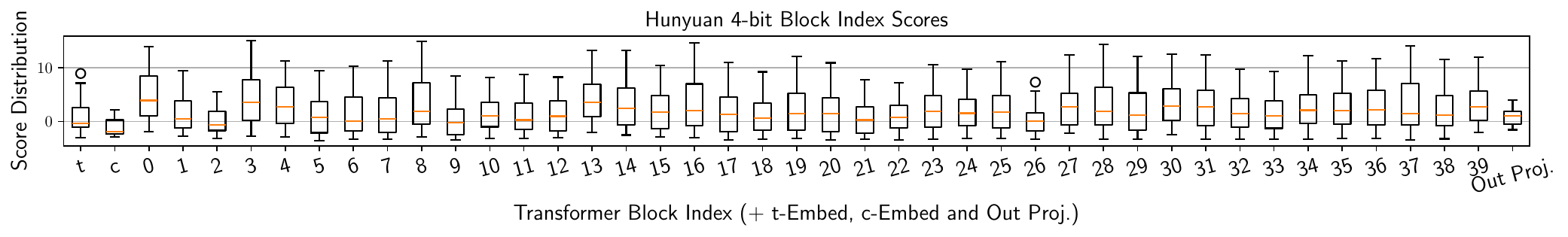}
    \includegraphics[height=1.05in]{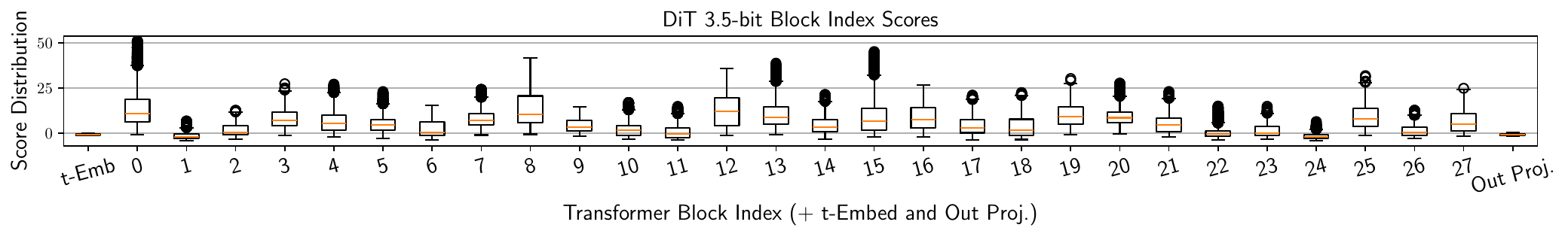}
    \includegraphics[height=1.05in]{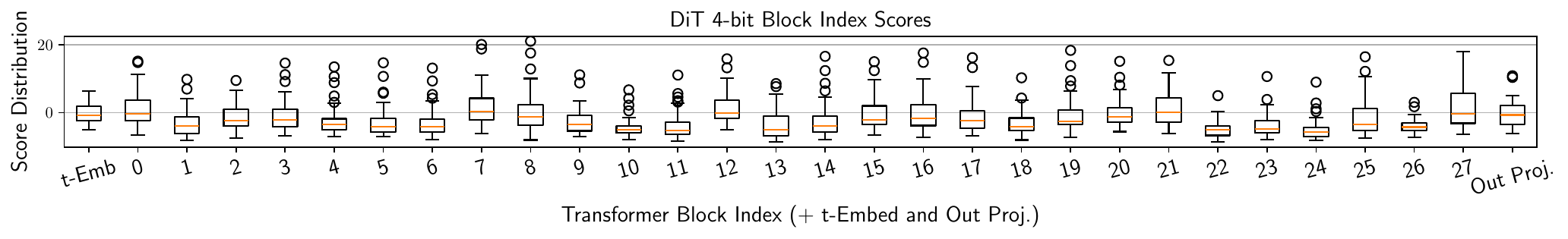}
    \caption{Layer-wise score box plots for several DiT models at multiple weight bit-precision levels.}
    \label{fig:dit_layerwise}
\end{figure*}

\subsubsection{Diffusion Transformer Layer-wise Sensitivity. } Figure~\ref{fig:dit_layerwise} provides layer-wise scores for both PixArt DMs, Hunyuan-DiT and DiT-XL/2 at different weight quantization bit precision levels. The purpose of these plots is to profile the importance of the sequential, yet mostly identical transformer blocks across the depth of each denoiser. Broadly speaking, for each DM and bit precision we observe that score distributions follow a sinusoidal wave pattern across network depth. This is a curious finding that may be useful in the future, e.g., applying LLM block pruning techniques \cite{gromov2024unreasonable}. Next, note how in every case the scores assigned to time-step embedding, i.e., `t-Emb' or `t' tend to outweight those for prompt/context embedding `c-Emb' or `c'. 

Additionally, we note that the scores assigned for the output layers `Out Proj.' have a large variance (the exception being DiT-XL/2 3.5-bit), indicating the certain methods of quantizing those layers can contribute to adequate or inadequate model performance. Finally, note the lopsided importance of the latter transformer layers (indices 21 and above) for the Hunyuan-DiT 3.65-bit quantization configuration. These blocks contain linear layers that interface with long residual skip-connections similar to what U-Nets have, but which can be quite sensitive to quantization per Fig.~\ref{fig:hunyuan_graph_op_box_365} as this is where the `bimodal activation distribution' problem discovered by Q-Diffusion~\cite{li2023q} would manifest.

\subsubsection{U-Net 4-bit Block-level Sensitivity. } 
\begin{figure}
    \centering
    \includegraphics[height=1.165in]{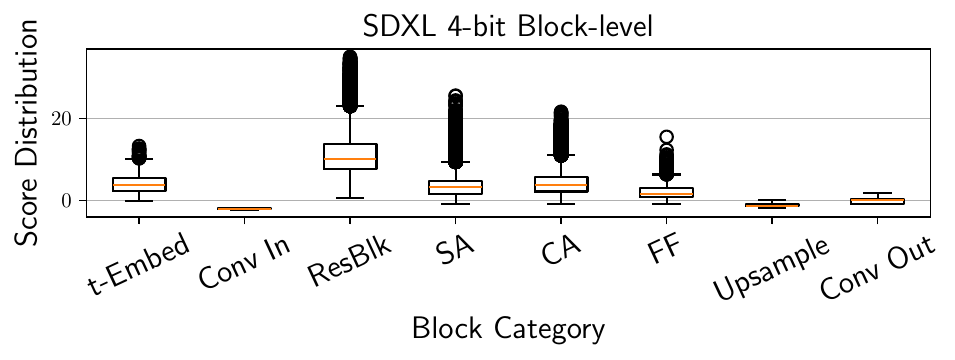}
    \qquad
    \includegraphics[height=1.165in]{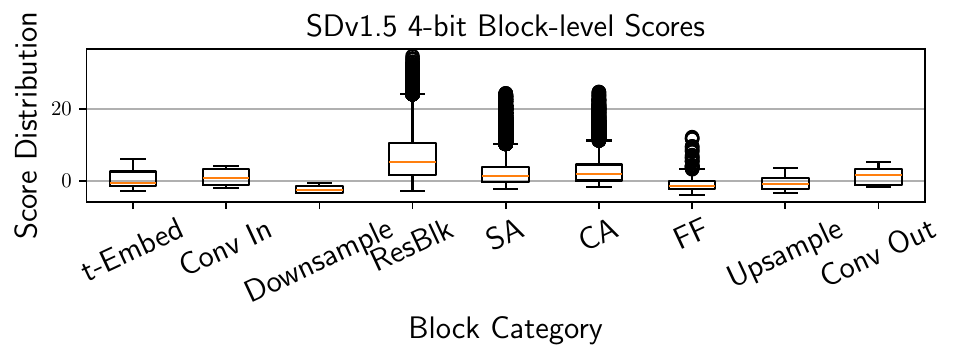}
    \caption{Box plot block-level score distributions for the 4-bit SDXL and SDv1.5 quantization configurations.}
    \label{fig:sdv15_box_sg_4bit}
\end{figure}

Figure~\ref{fig:sdv15_box_sg_4bit} provides block-wise quantization sensitivity score distributions for the SDXL and SDv1.5 U-Nets quantized to 4-bits. Note the extremely wide score distributions present for the ResNet blocks `ResBlk' category. Even though SDXL contains many more Transformer blocks than ResNet blocks~\cite{podell2023sdxl}, proper quantization of the former primarily controls the efficacy of the model after PTQ. Additionally, we also note that incremental importance of proper quantization for the self-attention and cross-attention structures compared to the feedforward component. Finally, the `Upsample' layers in SDXL do not seem as important as they are for SDv1.5, likely because the SDv1.5 feature pyramid contains one additional tier compared to SDXL.

\begin{figure}[t!]
    \centering
    \includegraphics[width=3.1in]{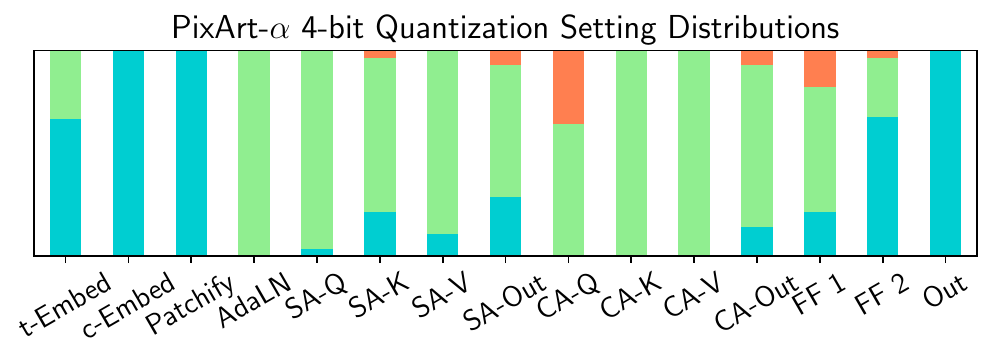}
    \includegraphics[width=3.1in]{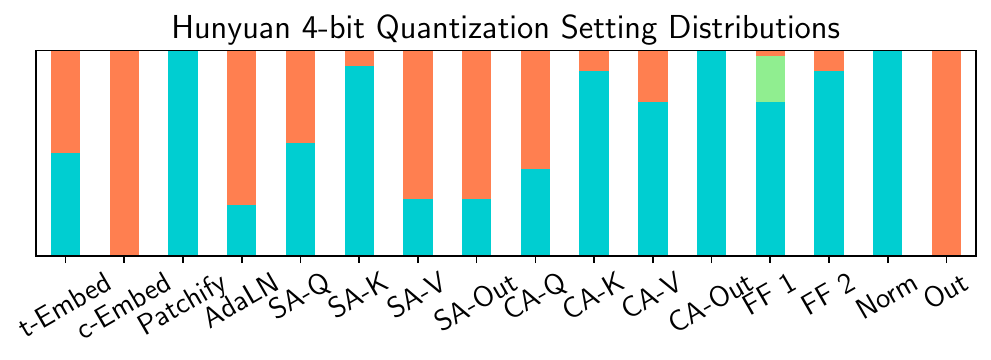}
    \includegraphics[width=3.1in]{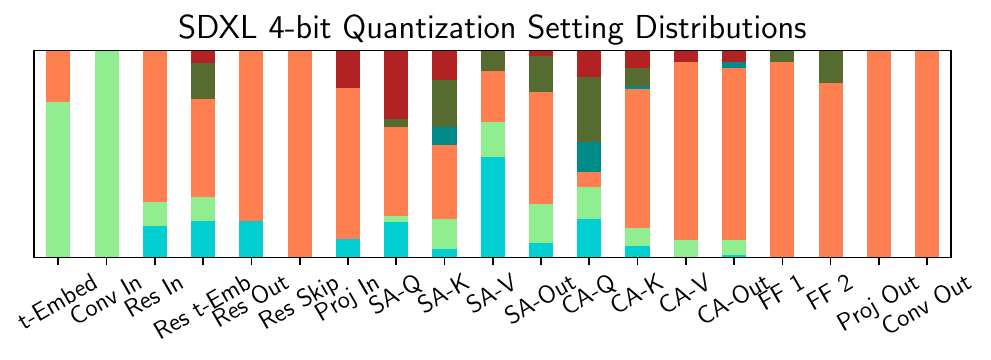}
    \includegraphics[width=3.1in]{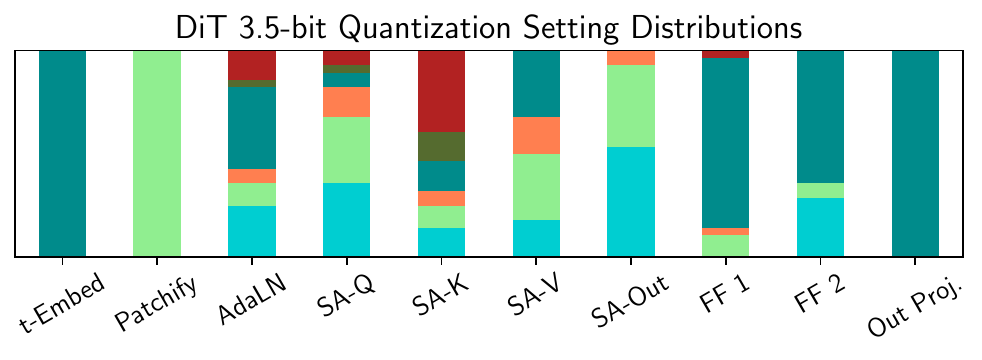}
    \includegraphics[width=3.1in]{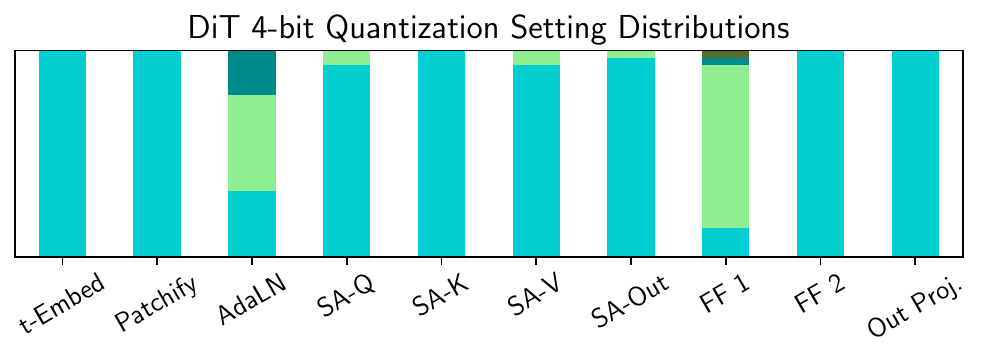}
    \includegraphics[width=3.2in]{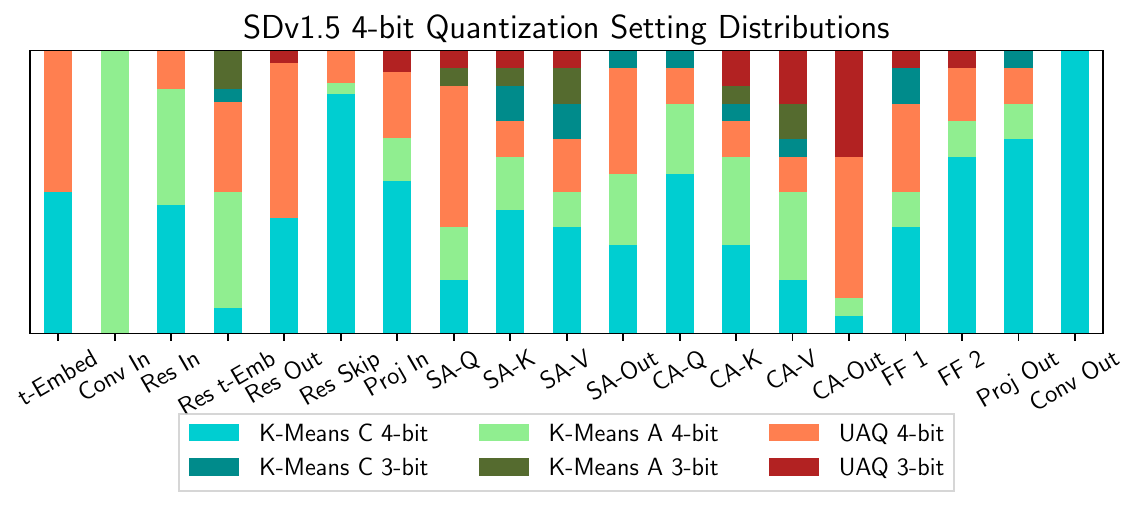}
    \caption{Stacked quantization method bar plots for remaining quantization configurations. Best viewed in color.}
    \label{fig:bar_plots_app}
\end{figure}

\subsubsection{Additional Quantization Setting Distributions. } Figure~\ref{fig:bar_plots_app} shows the additional quantization setting distributions for several DiT and U-Net DMs. Taken alongside Fig.~\ref{fig:bar_plots_main}, we observe that DiT denoisers generally prefer $K$-Means quantization to UAQ when the floating point overhead is not a concern. The exception is Hunyuan-DiT which at 4-bit precision tends to favor them equally depending on the operation category, while DiT-XL/2 heavily prefers cluster-based quantization even under a constrained optimization. By contrast, even when maximizing for pure performance SDXL features a strong preference for UAQ while SDv1.5 tends to mix and match quanitzation methods but shows a slight preference to channel-wise $K$-Means.

\begin{figure*}[t]
    \centering
    \subfloat[DiT $512^2$ resolution ImageNet images]{\includegraphics[width=3.4in]{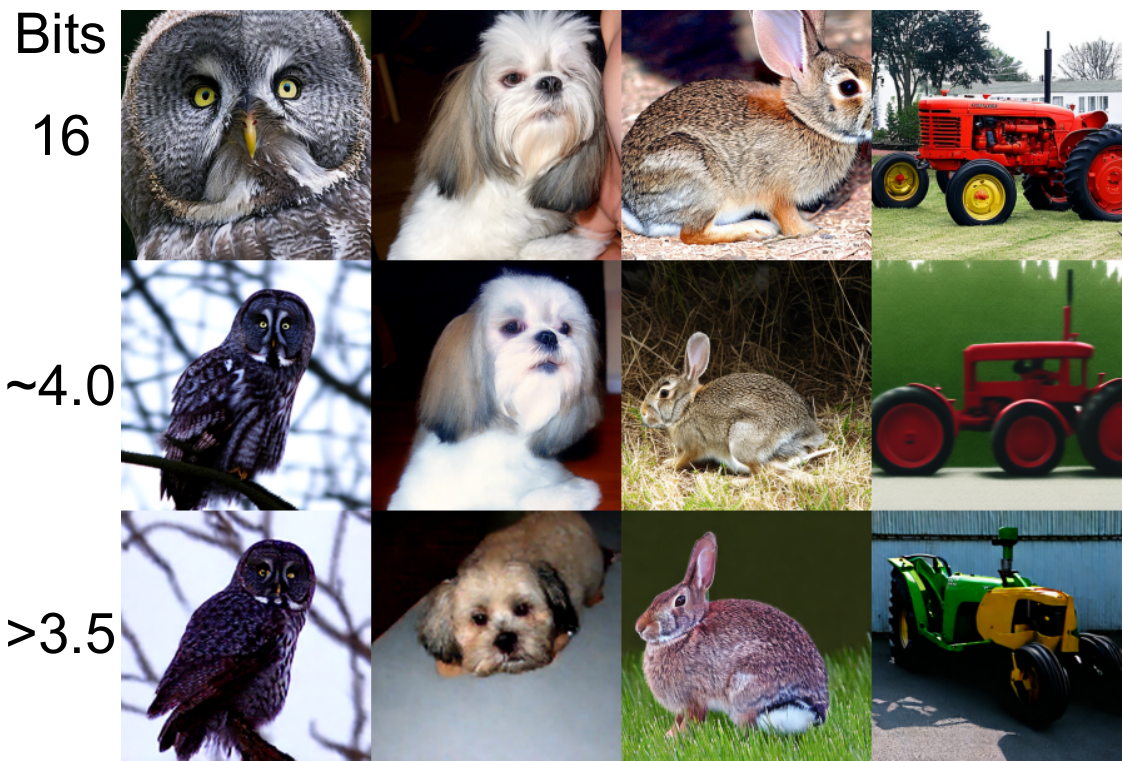}}
    \subfloat[SDXL $512^2$ resolution images w/ COCO prompts]{\includegraphics[width=3.4in]{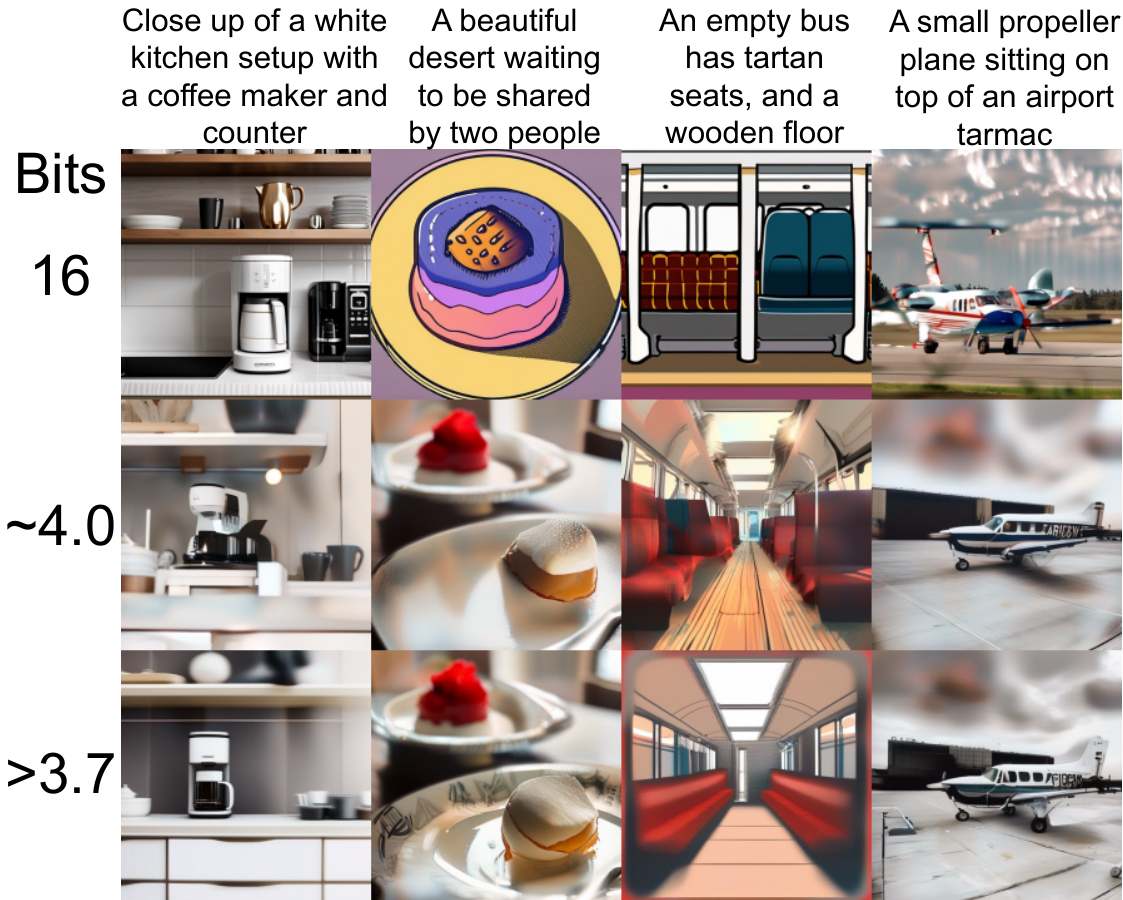}}
    \caption{Sample images by Qua$^2$SeDiMo quantized models compared to the original FP16 weights. (a) Generated by DiT-XL/2 and quantized either to to 4 (`NDCG Op-Level' predictor when $y=-FID$ in Fig.~\ref{fig:pareto_dit_v15}) or sub 3.5-bits (`NDCG Block-Level' predictor when $y=-FID-\lambda\widebar{Bits}$ in Fig.~\ref{fig:pareto_dit_v15}). (b) Generated by SDXL to 4 or sub 3.7-bits (`Hybrid Block-Level' predictors from FID and constrained optimization from Fig~\ref{fig:pareto_alpha_sigma_hunyuan_xl_fid} and Fig.~\ref{fig:pareto_dit_sdxl}, respectively).}
    \label{fig:sample_images}
\end{figure*}

\subsection{DiT-XL/2 and U-Net Visual Examples} Figure~\ref{fig:sample_images} provides sample images generated by DiT and SDXL models quantized using Qua$^2$SeDiMo compared to ones from the FP16 version. Note how most of these images contain a degree of realistic detail, e.g., the dog and rabbit generated by DiT. For the tractor image, the 4-bit model maintains the color and orientation while the sub 3.5-bit model has better details. For SDXL, images maintain a fair degree of detail, e.g., the `white kitchen' prompt. Sometimes, content generated by the quantized model are more realistic, e.g., for the `beautiful desert' and `empty bus' prompts.

\begin{figure}[t]
    \centering
    \includegraphics[width=3.35in]{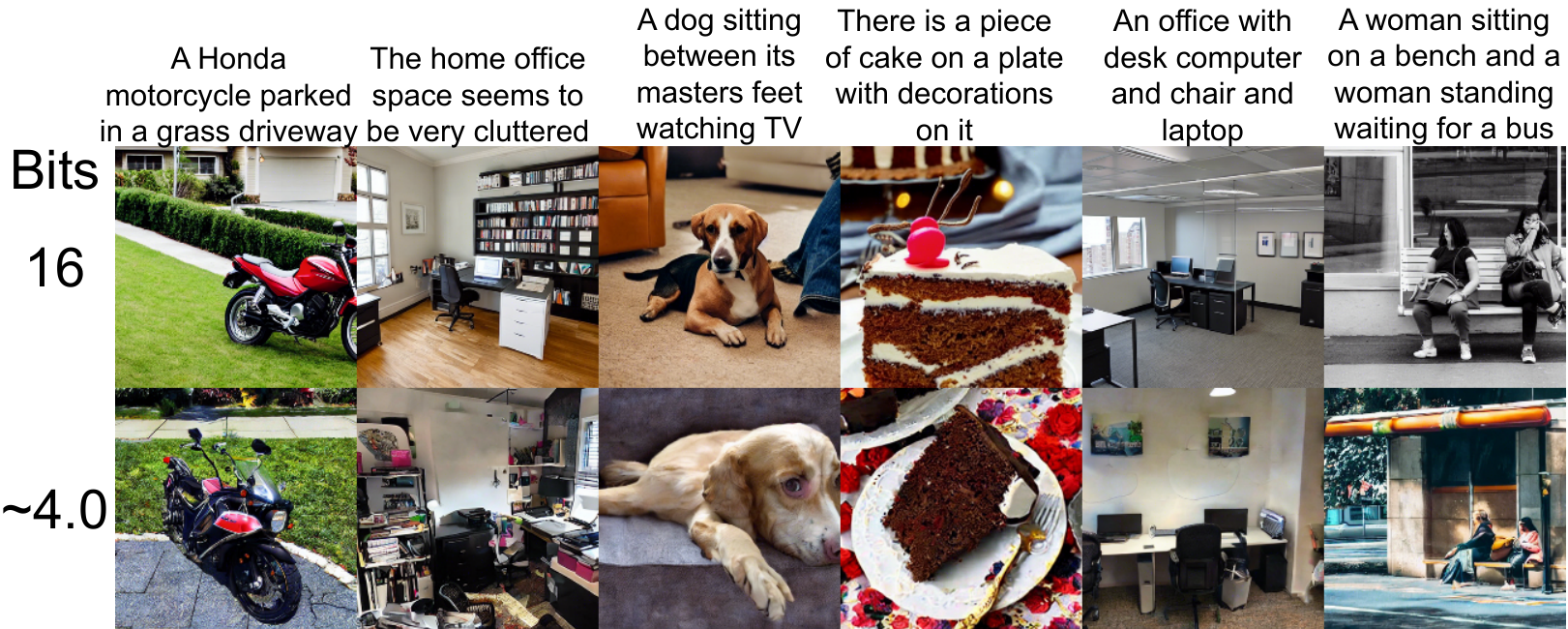}
    \qquad
    \caption{Annotated images generated by SDv1.5 with FP16 weights and quantized by Qua$^2$SeDiMo to 4-bits (using `Hybrid Op-Level' predictors for FID optimization in Fig.~\ref{fig:pareto_dit_v15}).}
    \label{fig:examples_sdv15}
\end{figure}

Next, Figure~\ref{fig:examples_sdv15} provides sample images from one of our SDv1.5 quantization configurations (`NDCG Op-Level' when optimizing $y=-FID$) in contrast to those produced by the original FP denoiser. Note how our images maintain similar visual quality, in fact, we are able to catch some prompt details not present in the FP image. For example, the second prompt states `The home office seems to be very cluttered', which better describes the image from the 4-bit model, while the final image prompt did not specify a black and white picture, yet the FP model produced one regardless.

\subsection{Predictor Training Setup and Performance}
\label{app:predictor}

Qua$^2$SeDiMo predictors consist of an initial embedding layer, 4 message passing layers, and an output MLP. The initial embedding layer applies embedding layers \texttt{t.nn.Embedding} to categorical features like quantization method, bit precision, and position encodings (e.g., block index) before concatenating all features together before applying a sequence of \texttt{t.nn.Linear}, \texttt{t.nn.BatchNorm1d} and \texttt{t.nn.ReLU} operations. This forms the 0-hop embedding for each individual node which is utilized for Op-level optimization. Message passing GNN layers have a hidden size of 64. Each GNN layer consists of a PyTorch-Geometric~\cite{Fey/Lenssen/2019} GATv2~\cite{brody2022how} module followed by a \texttt{t.nn.BatchNorm1d} operation and \texttt{t.nn.ReLU} activation. A residual connection links the input to the output. Finally, the MLP head consist of four \texttt{t.nn.Linear} with one \texttt{t.nn.ReLU} in the middle. 

We apply the $K=5$-fold predictor ensemble scheme from Sec.~\ref{sec:dataset_stats}. Each predictor trains for 10k epochs using a batch size of 128. We use the AdamW~\cite{loshchilov19AdamW} optimizer with initial learning rate of $10^{-3}$ and L2 weight decay of $10^{-6}$. Additionally, we anneal the learning rate via cosine scheduler~\cite{Loshchilov2017SGDRSG}. Also, depending on denoiser architecture search space, we modify Eq.~\ref{eq:ab_loss} to only perform for hop-levels that contain subgraphs we will use to construct quantization configurations, see Sec.~\ref{app:denoiser_subgraphs} for further details. We train all predictors in the ensemble simultaneously using multi-threading. It takes about an hour and 4-10GB of VRAM to train and evaluate an ensemble of 5 predictors. 

\begin{figure*}[t!]
    \centering
    \subfloat{\includegraphics[width=6.85in]{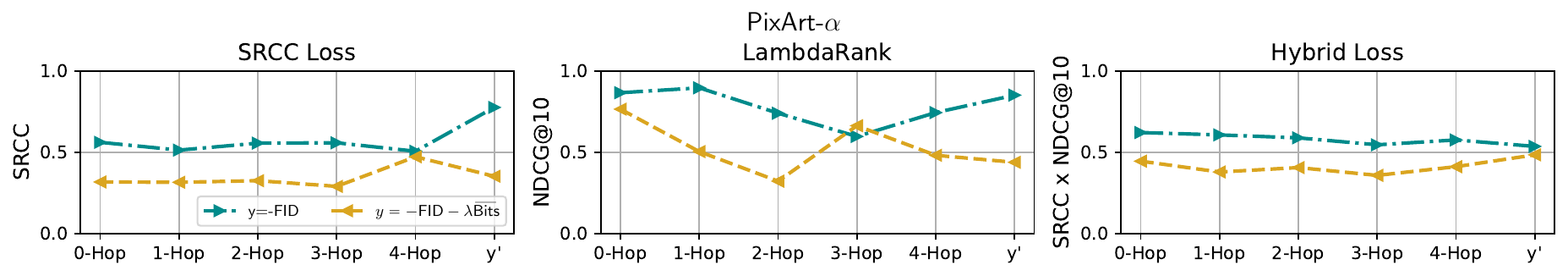}}
    \qquad
    \subfloat{\includegraphics[width=6.85in]{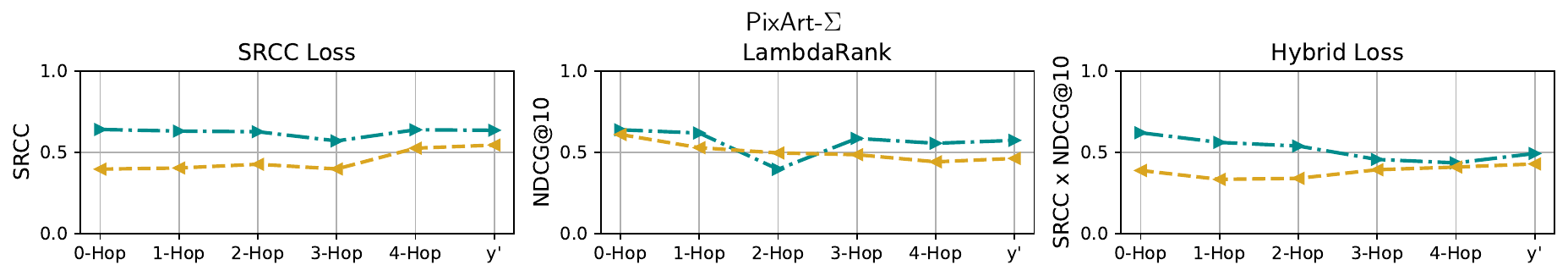}}
    \qquad
    \subfloat{\includegraphics[width=6.85in]{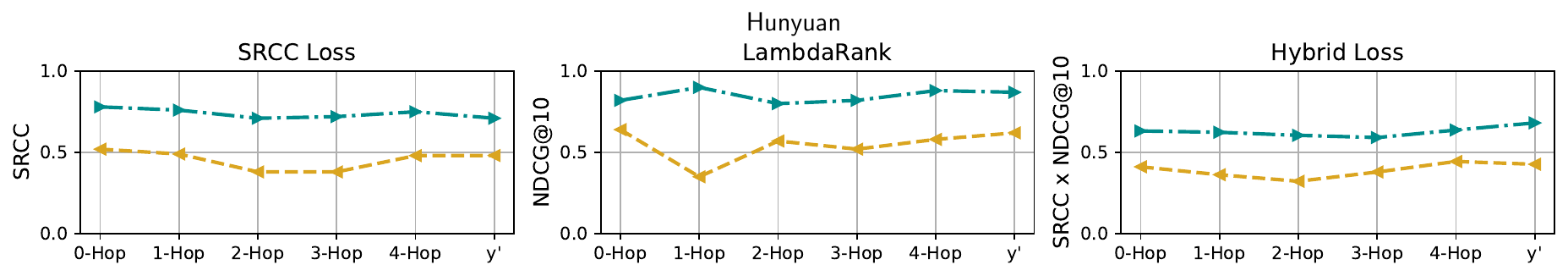}}
    \qquad
    \subfloat{\includegraphics[width=6.85in]{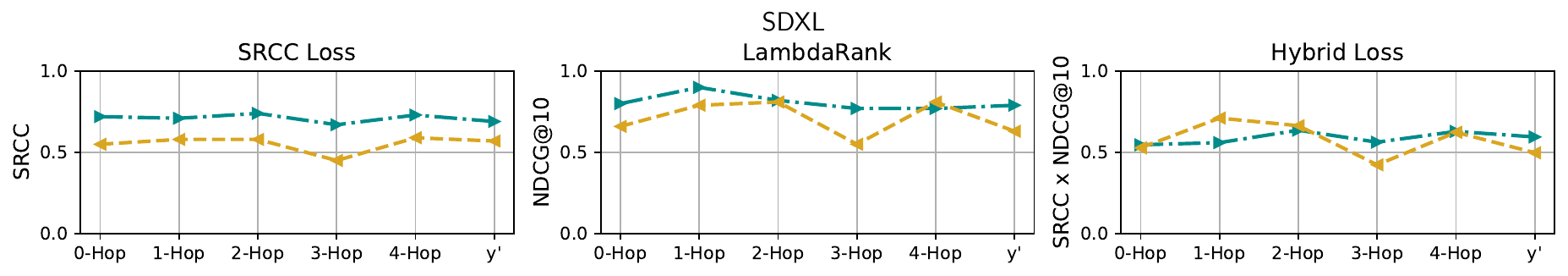}}
    \qquad
    \subfloat{\includegraphics[width=6.85in]{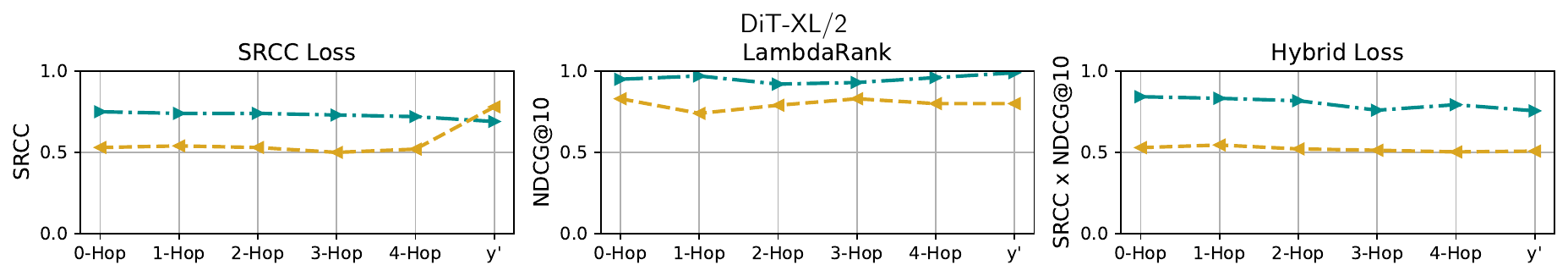}}
    \qquad
    \subfloat{\includegraphics[width=6.85in]{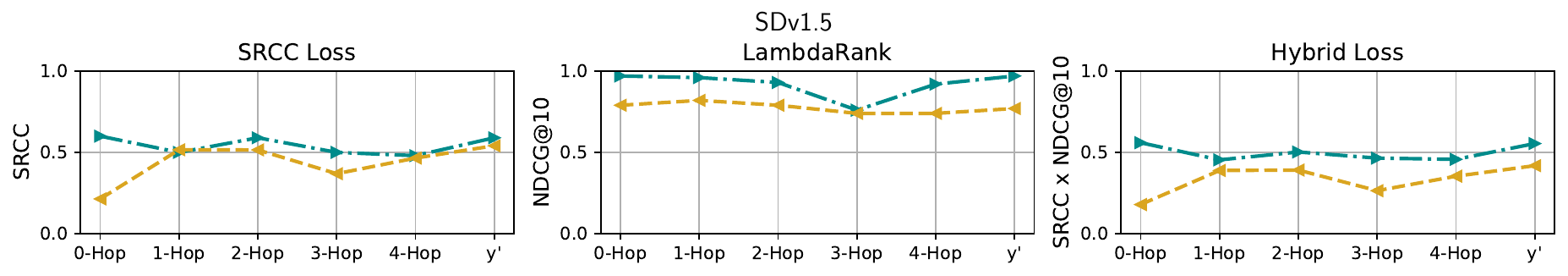}}
    \caption{Validation set predictor performance across three types of hop-level losses $\mathcal{L}_{rank}$ and two target equations. X-Axis refers to the performance of embedding norms at different hop levels and the MLP regression head. Results averaged across $K=5$ folds.}
    \label{fig:predictor_performance}
\end{figure*}

After predictor training, we standardize the scores for each hop-level into a normal distribution $\mathcal{N}(0, 1)$ using statistics from the training data. We then calculate ensemble weights using validation set performance (e.g., SRCC, NDCG@10, or their product for `Hybrid' $\mathcal{L}_{rank}$) for each hop-level, then enumerate and score all possible subgraphs. Specifically, we do two rounds of scoring: First is Op-level, where we consider each weight layer node individually and there are $\#W$ nodes to consider, and $6$ quantization settings per node, so $6\times{\#W}$ scores generated. Second, for Block-level, we look at the subgraph-level scores. For each subgraph, there are $6^{\#W_{SG}}$ subgraphs to consider where $\#W_{SG}$ is the number of weight layer nodes in the subgraph. 

Finally, Figure~\ref{fig:predictor_performance} illustrates the predictor performance on the validation set. Note that the performance metrics here correspond to ensemble weights when computing subgraph scores. Generally the predictors obtain adequate validation performance above 0.5 SRCC or NDCG@10 per hop-level, but they have an easier time maximizing the latter. We observe that it is typically harder to achieve high SRCC/NDCG when optimizing $y=-FID-\lambda\widebar{Bits}$, but this is also denoiser-dependant. Sometimes the hop-level ranking performance exceeds that of the MLP head, but this is because the MLP head is not trained using a ranking loss, but as a regressor using mean-squared-error.

\subsection{Denoiser Subgraphs}
\label{app:denoiser_subgraphs}

Our Block-level profiling scheme using subgraphs presents two design choices: One, since the GNN consists of multiple layers and each layer providing an embedding, which ones should we extract quantifiable insights from? Two, since the same node can be part of the multiple induced subgraphs for the same value of $m$, which should we use to guide the construction of new quantization configurations? First, we consider two approaches for constructing optimal quantization configurations. The simpler approach only considers the $m=0$ embedding for each node and optimizes them in isolation. However, this may not be optimal as the weights of a neural network do not exist in isolation, necessitating the need for subgraphs that capture neighborhood information.

Second, to facilitate subgraph-level optimization, Qua$^2$SeDiMo employs a greedy heuristic approach. We carefully determine the largest appropriate subgraph partitions that encompass distinct components like ResNet blocks, or attention modules, as shown in Fig.~\ref{fig:subgraphs}. However, it is crucial to plan this partitioning carefully, as GNN message passing may introduce information from undesired nodes depending on the choice of subgraph root. For instance, looking at Fig.~\ref{fig:subgraphs}, if we use a 4-hop subgraph rooted at the `Add' node to represent DiT attention, we would include unnecessary information from the nodes that feed into `x-input' while missing information from the AdaLN `Linear' node. This issue does not arise when `Proj Out' is the root.

One implication of this approach is that we only require a subset of the embeddings at a given hop level to compute the subgraph scores. For example, in Fig.~\ref{fig:subgraphs}, since `FF 2' is the root for the feedforward module, we compute the score as $\norm{h_{FF 2}^1}_1$, while $\norm{h_{FF 1}^1}_1$ is unused. Formally, we define $\mathcal{V}_\mathcal{G}^m \in \mathcal{V}_\mathcal{G}$ as the subset of nodes corresponding to the subgraph roots at a given hop-level that we use to construct a quantization configuration. Further, we augment the graph aggregation Eq.~\ref{eq:aggr} to enumerate over $\mathcal{V}_\mathcal{G}^m$ rather than $\mathcal{V}_\mathcal{G}$. This selectively limits the embeddings involved when calculating $\mathcal{L}_{rank}$, compelling the GNN to focus on accurately scoring the subgraphs we draw insights and use to generate optimal quantization configurations. However, these two design choices necessitate careful design of the Block-level subgraphs we consider. We now provide extensive details on this matter below:

To enumerate the selected Block-level subgraphs we use when optimizing each denoiser architecture, we mention the number of weight layer nodes, hop-level and which node serves as the root. The scripts for all subgraphs are located in the code submission. 
\subsubsection{PixArt-$\alpha$/$\Sigma$} each contain a total of 39576 possible subgraphs split between 8 categories. 
\begin{enumerate}
    \item AdaLN/Time Embedding: 2-hop subgraph covering the three time-embedding linear layers.
    \item Caption-Embedding: 1-hop subgraph for the two linear layers and rooted at the second layer.
    \item Patchify: 0-hop subgraph containing the patch-embedding convolution operation weight layer.
    \item Self-Attention: 3-hop subgraph containing 7 nodes, 4 of which are weight layers: Q, K, V and output projection layer. Rooted at the output projection layer.
    \item Cross-Attention (1): 1-hop subgraph containing the Q and K weight layers and a dummy `MatMul' node for the QK product. Rooted at the `MatMul' node.
    \item Cross-Attention (2): 2-hop subgraph containing the V and output projection weight layers, and rooted at the latter.
    1-hop subgraph containing 3-nodes: 
    \item Feedforward: 1-hop subgraph for the two linear layers. Root is the second layer.
    \item Projection Out: 1-hop subgraph containing the final `norm\_out' and `proj\_out' layers and rooted at the latter.
\end{enumerate}

\subsubsection{Hunyuan-DiT} contains a total of 317418 possible subgraphs split between 9 categories. 
\begin{enumerate}
    \item AdaLN/Time Embedding: 2-hop subgraph for the three time-step weight layers. Root is the final linear layer.
    \item Caption-Embedding: 1-hop subgraph for the two linear layers and rooted at the second layer.
    \item Patchify: 0-hop subgraph containing the patch-embedding convolution operation weight layer.
    \item Self-Attention: 3-hop subgraph containing 7 nodes, 4 of which are weight layers: Q, K, V and output projection layer. Rooted at the output projection layer.
    \item Cross-Attention (1): 1-hop subgraph containing the Q and K weight layers and a dummy `MatMul' node for the QK product. Rooted at the `MatMul' node.
    \item Cross-Attention (2): 2-hop subgraph containing the V and output projection weight layers, and rooted at the latter.
    \item Feedforward: 1-hop subgraph for the two linear layers. Root is the second layer.
    \item Skip-connection: 1-hop subgraph rooted at an `Add' node that sums the output of the previous transformer block with the input from the long residual skip-connection.
    \item Projection Out: 0-hop subgraph containing the final `proj\_out' weight layer in the DiT.
\end{enumerate}

\subsubsection{SDXL} contains a total of 343458 possible subgraphs split between 11 categories. 
\begin{enumerate}
    \item Time-Embedding: 1-hop subgraph for the two time-step weight layers. Root is the second linear layer.
    \item Input Convolution: 0-hop subgraph consisting of the first convolution in the U-Net.
    \item Output Convolution: 0-hop subgraph consisting of the last convolution in the U-Net.
    \item Upsampler: 0-hop subgraph consisting of an upsampling conv.
    \item Self-Attention: 3-hop subgraph consisting of 5 weight layers: projection in convolution, Q, K, V and output projection linear layers. Rooted at the output projection layer.
    \item Cross-Attention: 3-hop subgraph consisting of 4 weight layers: Q, K, V and output projection linear layers. Rooted at the output projection layer. Note: Where self-attention contains an input projection convolution, cross-attention contains a dummy `Add' node from the self-attention that has non-positional features set to 0.
    \item Feedforward: 1-hop subgraph for the two linear layers. Root is the second layer.
    \item Transformer Projection Out: 0-hop subgraph consisting of the output projection layer following the feedforward.
    \item Input ResNet Block w/out Skip: 1-hop subgraph consisting of the input, output and time-embedding layers. Rooted at the output layer.
    \item Input ResNet Block w/skip: 2-hop subgraph consisting of 5 weight layers: input, output, time-embed, skip-connection and a downsampling layer. Rooted at a dummy `Add' node.
    \item Output ResNet Block: 2-hop subgraph consisting of 5 weight layers: input, output, time-embed and two skip-connection layers following~\cite{li2023q}. Rooted at an output dummy `Add' node.
\end{enumerate}

\subsubsection{DiT} contains a total of 257688 possible subgraphs split between 4 categories. 
\begin{enumerate}
    \item Time-Embedding: 1-hop subgraph for the two time-step weight layers. Root is the second linear layer.
    \item Attention: 4-hop subgraph containing 8 nodes, at least 5 are weight layers: AdaLN linear layer, Q, K, V and output projection layer. Attention of the first DiT block contains the `Patchify' convolution. Rooted at the output projection layer.
    \item Feedforward: 1-hop subgraph for the two linear layers. Root is the second layer.
    \item Projection Out: 1-hop subgraph for the final two convolution layers in the DiT, after the final Transformer block. Root is the second conv layer.
\end{enumerate}

\subsubsection{SDv1.5} contains a total of 256488 possible subgraphs split between 12 categories. 
\begin{enumerate}
    \item Time-Embedding: 1-hop subgraph for the two time-step weight layers. Root is the second linear layer.
    \item Input Convolution: 0-hop subgraph consisting of the first convolution in the U-Net.
    \item Output Convolution: 0-hop subgraph consisting of the last convolution in the U-Net.
    \item Block 9 Downsampler: 0-hop subgraph consisting of one of the downsampling convolutions. The other downsampling convolutions are merged into ResNet block subgraphs.
    \item Upsampler: 0-hop subgraph consisting of an upsampling conv.
    \item Self-Attention: 3-hop subgraph consisting of 5 weight layers: projection in convolution, Q, K, V and output projection linear layers. Rooted at the output projection layer.
    \item Cross-Attention: 3-hop subgraph consisting of 4 weight layers: Q, K, V and output projection linear layers. Rooted at the output projection layer. Note: Where self-attention contains an input projection convolution, cross-attention contains a dummy `Add' node from the self-attention that has non-positional features set to 0.
    \item Feedforward: 1-hop subgraph for the two linear layers. Root is the second layer.
    \item Transformer Projection Out: 0-hop subgraph consisting of the output projection layer following the feedforward.
    \item Input ResNet Block w/out Skip: 1-hop subgraph consisting of the input, output and time-embedding layers. Rooted at the output layer.
    \item Input ResNet Block w/skip: 2-hop subgraph consisting of 4 or 5 weight layers: input, output, time-embed, and skip-connection layers. 5th layer is a downsampling convolution. Rooted at a dummy `Add' node.
    \item Output ResNet Block: 2-hop subgraph consisting of 5 weight layers: input, output, time-embed and two skip-connection layers following~\cite{li2023q}. Rooted at an output dummy `Add' node.
\end{enumerate}

\subsection{Experimental Hardware and Software Resources}
\label{app:hw_setup}

All experiments conducted in this paper were performed on a rack server with 8 NVIDIA V100 32GB GPUs, an Intel Xeon Gold 6140 GPU and 756GB RAM. We run our experiments in Python 3 using two different anaconda virtual environments and open-source repository forks: 
\begin{itemize}
    \item All Diffusion Model experiments, e.g., sampling and evaluating quantization configurations, generating images, evaluating FID, etc., stem from a fork of the Q-Diffusion~\cite{li2023q} repository. We modify the virtual environment to use more up-to-date versions of some packages that support newer Diffusion Models, e.g., Hunyuan-DiT. Please see the README in the code submission for details. 
    \item All predictor experiments, e.g., training, generating optimal quantization configurations, stem from a fork of the AutoBuild~\cite{mills2024autobuild} repository. 
\end{itemize}